\def\paperlanguage{}
\pdfoutput=1 % for arxiv

\documentclass[letterpaper, 10 pt, conference]{ieeeconf}  % Comment this line out if you need a4paper

\usepackage{bm}
\usepackage{cite}
\usepackage{flushend}
\include{preamble}

\IEEEoverridecommandlockouts                              % This command is only needed if
\title{\LARGE \textbf
  {
    \switchlanguage%
    {%
      Musculoskeletal AutoEncoder:\\A Unified Online Acquisition Method of Intersensory Networks for State Estimation, Control, and Simulation of Musculoskeletal Humanoids
    }%
    {%
      Musculoskeletal AutoEncoder:\\筋骨格ヒューマノイドの状態推定・制御・シミュレーションを\\統一的に扱う筋骨格センサ間ネットワークのオンライン獲得手法
    }%
  }
}

\author{Kento Kawaharazuka$^1$, Kei Tsuzuki$^1$, Moritaka Onitsuka$^1$, Yuki Asano$^1$\\Kei Okada$^1$, Koji Kawasaki$^2$, and Masayuki Inaba$^1$% <-this % stops a space
  \thanks{$^1$ The author are associated with Department of Mechano-Informatics, Graduate School of Information Science and Technology, The University of Tokyo. %, 7-3-1 Hongo, Bunkyo-ku, Tokyo, 113-8656, Japan.
    \texttt\small \{kawaharazuka, tsuzuki, onitsuka, asano, k-okada, inaba\}@jsk.t.u-tokyo.ac.jp
  }
  \thanks{$^{2}$ The author is associated with TOYOTA MOTOR CORPORATION.
    {\texttt\small koji\_kawasaki@mail.toyota.co.jp}
    }
}
\begin{document}

\maketitle
\thispagestyle{empty}
\pagestyle{empty}

%%%%%%%%%%%%%%%%%%%%%%%%%%%%%%%%%%%%%%%%%%%%%%%%%%%%%%%%%%%%%%%%%%%%%%%%%%%%%%%%
\begin{abstract}
  \switchlanguage%
  {%
    While the musculoskeletal humanoid has various biomimetic benefits, the modeling of its complex structure is difficult, and many learning-based systems have been developed so far.
    There are various methods, such as control methods using acquired relationships between joints and muscles represented by a data table or neural network, and state estimation methods using Extended Kalman Filter or table search.
    In this study, we construct a Musculoskeletal AutoEncoder representing the relationship among joint angles, muscle tensions, and muscle lengths, and propose a unified method of state estimation, control, and simulation of musculoskeletal humanoids using it.
    By updating the Musculoskeletal AutoEncoder online using the actual robot sensor information, we can continuously conduct more accurate state estimation, control, and simulation than before the online learning.
    We conducted several experiments using the musculoskeletal humanoid Musashi, and verified the effectiveness of this study.
  }%
  {%
    筋骨格ヒューマノイドは様々な生物規範型の利点を有すると同時にそのモデリングは困難であり, これまで多くの学習制御手法が開発されてきた.
    制御においてはニューラルネットワークやデータテーブルを用いた関節と筋の関係の学習, 関節角度推定においては拡張カルマンフィルタを用いた手法等様々である.
    本研究では, AutoEncoder型の関節角度・筋張力・筋長の関係を表すニューラルネットワークを構築することで, 筋骨格ヒューマノイドの状態推定・制御・シミュレーションを統一的に行う手法を提案する.
    また, 本ネットワークを実機センサ情報からオンラインで更新することで, より正確な状態推定・制御・シミュレーションが継続的に可能となる.
    筋骨格ヒューマノイドMusashiを用いて実機において検証を行い, 本手法の有効性を確認した.
  }%
\end{abstract}

\section{INTRODUCTION}\label{sec:introduction}
\begin{figure}[t]
  \centering
  \includegraphics[width=1.0\columnwidth]{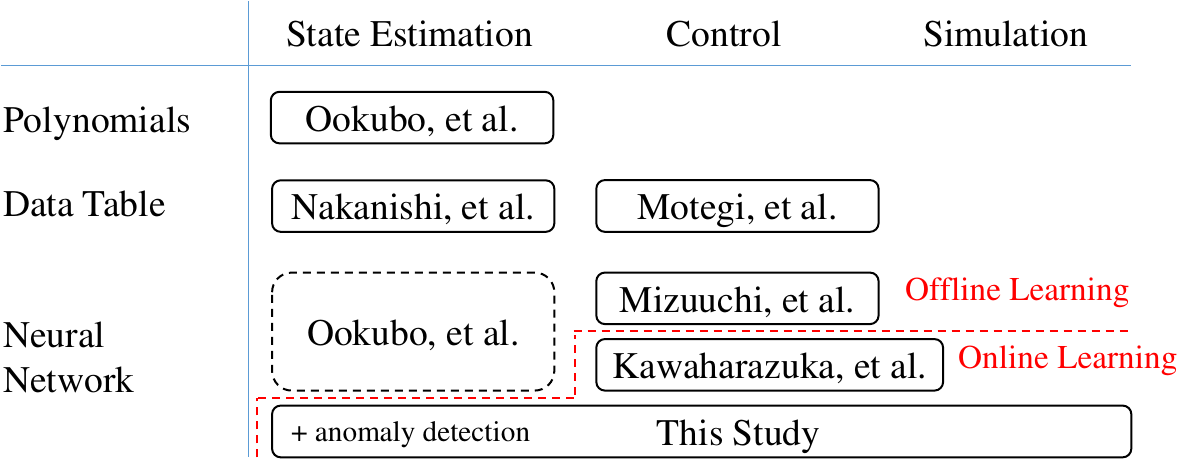}
  \vspace{-3.0ex}
  \caption{Classification of previous studies and positioning of this study.}
  \label{figure:previous-studies}
  \vspace{-3.0ex}
\end{figure}

\switchlanguage%
{%
  The musculoskeletal humanoid \cite{nakanishi2013design, wittmeier2013toward, jantsch2013anthrob, asano2016kengoro} has many biomimetic benefits such as muscle redundancy, variable stiffness control with nonlinear elastic elements, joint torque control with cheap muscle tension sensors, and the flexible under-actuated fingers and spine.
  On the other hand, its complex musculoskeletal structure is difficult to modelize, and applying conventional control methods is challenging.

  To solve this problem, many learning-based systems have been developed so far.
  In this study, we mainly handle the state estimation, control, and simulation methods of musculoskeletal humanoids.
  The state estimation is to estimate joint angle from muscle length and muscle tension, the control is to calculate target muscle length from target joint angle and torque, and the simulation is to simulate the transition of joint angle and muscle tension from target muscle length and external force.
  We summarize these studies that especially handle the actual robot in \figref{figure:previous-studies}.

  First, because the musculoskeletal humanoid typically does not have joint angle sensors due to complex joints, such as the scapula and shoulder ball joints, several methods have been developed to estimate joint angles.
  Ookubo, et al. have trained a nonlinear relationship between joint angles and muscle lengths (joint-muscle mapping, JMM), represented by polynomials, using inertial measurement sensors (IMU), and estimated joint angles using the polynomials and Extended Kalman Filter (EKF) \cite{ookubo2015learning}.
  Nakanishi, et al. have constructed a data table representing JMM, and estimated the joint angles using muscle length sensors and table search algorithm \cite{nakanishi2010estimation}.
  JMMs used in these methods are trained offline.
  Also, the method proposed by Ookubo, et al. can be applied to JMM represented by a neural network.

  Second, we will explain several learning-based control methods.
  Motegi, et al. have constructed a data table associating the position of end effectors and muscle lengths, and proposed a method to realize the accurate position of the end effectors \cite{motegi2012jacobian}.
  Mizuuchi, et al. have controlled the robot by training JMM represented by a neural network using motion capture data \cite{mizuuchi2006acquisition}.
  Kawaharazuka, et al. have developed a control method with JMM updated online using the actual robot sensor information \cite{kawaharazuka2018online}, and developed an extended control method considering the musculoskeletal body tissue softness \cite{kawaharazuka2018bodyimage, kawaharazuka2019longtime}.
  Although there are several control methods using reinforcement learning \cite{diamond2014reaching}, they are difficult to use for the actual robot.
  % There are several joint torque control methods \cite{jantsch2011scalable, jantsch2012computed, kawamura2016jointspace}, but the results have been poor regarding the actual robot experiments due to the friction of the musculoskeletal structure.
  Also, there are no studies applying learning-based methods to not only state estimation and control but also simulation of the musculoskeletal structure.
  We need to make the simulator closer to the actual robot by learning.
  On the other hand, there are several simulations using geometric models \cite{wittmeier2011caliper, lau2016caspr, wittmeier2011cardsflow}.
}%
{%
  筋骨格ヒューマノイド\cite{nakanishi2013design, wittmeier2013toward, jantsch2013anthrob, asano2016kengoro}は冗長な筋配置や非線形弾性要素による可変剛性制御, 安価な筋張力センサによるトルク制御や背骨等の劣駆動構造等, 多くの生物規範型の利点を有する.
  一方で, その複雑な筋骨格構造のモデリングは困難であり, 既存の制御手法を適用することが難しい.

  そのため, これまで多くの筋骨格構造に関わる学習型の制御手法や状態推定手法が開発されてきた.
  それらの研究の中でも特に実機を扱ったものを\figref{figure:previous-studies}にまとめる.
  まず状態推定について, 筋骨格ヒューマノイドは球関節等の存在から一般的には関節角度センサを有さなたいめ, 関節角度を推定する手法が開発されている.
  大久保らは, 慣性センサ等から得たデータを使って関節と筋の関係を多項式近似により表し, 拡張カルマンフィルタを用いて実機関節角度を推定している\cite{ookubo2015learning}.
  中西らは, 関節と筋の関係を表すデータテーブルを用いてマッチングを行うことで, 筋長センサから実機の関節角度を推定している\cite{nakanishi2010estimation}.
  大久保らの方法は, 関節と筋の関係をニューラルネットワークを用いて表現した場合にも適用することが可能である.
  次に制御手法について述べる.
  茂木らは, 手先位置と筋長を対応付けるテーブルを構築し, 正確な手先位置を実現する手法を提案している\cite{motegi2012jacobian}.
  水内らは, モーションキャプチャデータから関節と筋の関係をニューラルネットワークで表現し, 身体制御を行っている\cite{mizuuchi2006acquisition}.
  河原塚らは, 関節と筋の関係を表すニューラルネットワークをオンラインで学習し身体制御を行う手法\cite{kawaharazuka2018online}, さらにそれを拡張し, 筋張力の影響による身体組織の柔軟性を考慮した手法を開発している\cite{kawaharazuka2018bodyimage}.
  その他にも, 強化学習を用いる手法\cite{diamond2014reaching}があるが, 実機での使用は難しい.
  また, 筋張力による関節トルク制御手法\cite{jantsch2011scalable, jantsch2012computed, kawamura2016jointspace}も存在するが, これらは筋張ヤコビアンのみ必要なため本研究を用いても行うことができる.
  しかし, 摩擦等の関係から実機における適用においては, 良い結果は出ていない.
  また, 学習した結果を状態推定や制御に用いることはあっても, シミュレーションに対して適用した手法はない.
  補足として, 制作した幾何モデルを用いたシミュレーションはいくつか存在する\cite{wittmeier2011caliper, lau2016caspr, wittmeier2011cardsflow}.
}%

\begin{figure*}[t]
  \centering
  \includegraphics[width=2.0\columnwidth]{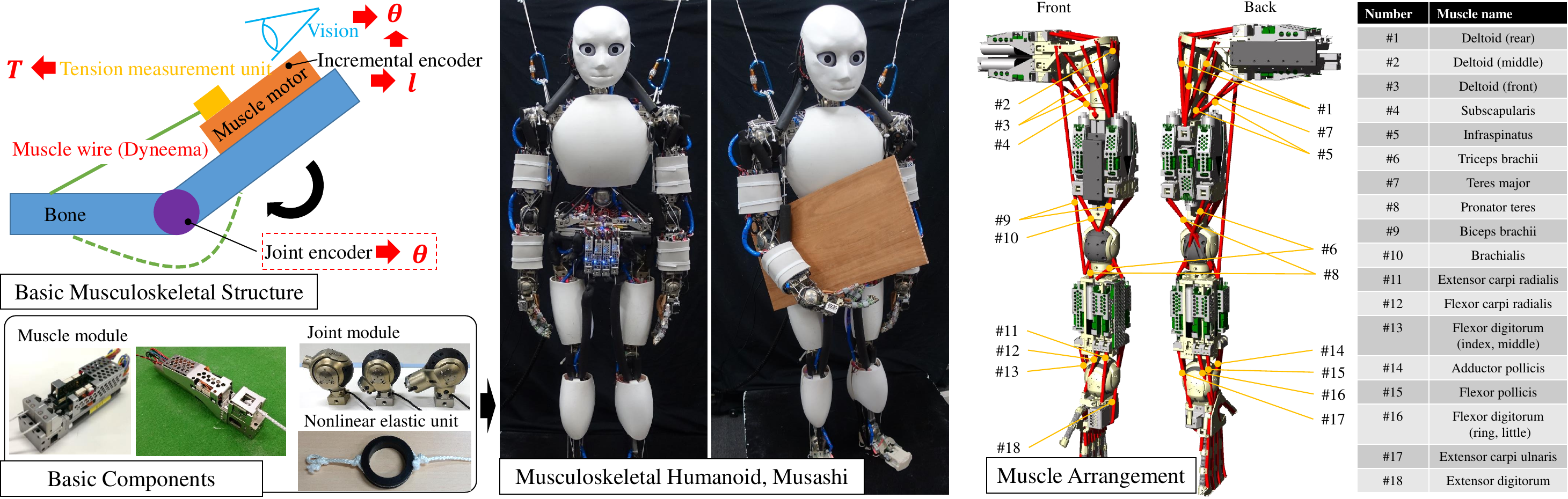}
  \caption{The basic structure of musculoskeletal humanoids covered in this study, and the musculoskeletal humanoid Musashi used in this study.}
  \label{figure:musculoskeletal-humanoid}
  \vspace{-3.0ex}
\end{figure*}

\switchlanguage%
{%
  In this study, we propose a Musculoskeletal AutoEncoder (MAE) uniformly handling the state estimation, control, and simulation of musculoskeletal humanoids.
  By updating MAE online using the actual robot sensor information, we can continuously conduct more accurate state estimation, control, and simulation of the musculoskeletal structure than before the online learning.
  This study only handles quasi-static states, and does not consider dynamic states.
  Also, we assume that the geometric model, including joint position, link length, and link weight, is correct.
  % In \secref{sec:musculoskeletal-introduction}, we will explain the basic structure of musculoskeletal humanoids and the relationship among musculoskeletal sensors.
  % In \secref{sec:musculoskeletal-autoencoder}, we will explain the structure of MAE and its initial training and online learning.
  % Also, we will explain state estimation, control, and simulation of the musculoskeletal structure using the acquired MAE.
  % In \secref{sec:experiments}, we will conduct experiments of state estimation, control, and simulation using the actual musculoskeletal humanoid Musashi, and verify the effectiveness of this study.
}%
{%
  そこで本研究では, 状態推定・制御・シミュレーションを統一的に扱うことができるMusculoskeletal AutoEncoderを提案する.
  本ネットワークを実機センサデータをもとにオンラインで更新していくことで, 筋骨格ヒューマノイドの状態推定・制御・シミュレーションをより正確かつ継続的に行うことができるようになる.
  また, 本研究では静的な関係のみを扱っているため, 動的な要素は考慮していない.
  \secref{sec:musculoskeletal-introduction}では, 筋骨格ヒューマノイドの基本構造とそのセンサ群の関係性について述べる.
  \secref{sec:musculoskeletal-autoencoder}では, 提案するMusculoskeletal AutoEncoderの構造とその初期学習・オンライン学習について述べる.
  また, Musculoskeletal AutoEncoderを用いた筋骨格ヒューマノイドの状態推定・身体制御・シミュレーション方法について述べる.
  \secref{sec:experiments}では, 本手法を用いた筋骨格ヒューマノイドMusashiの実機における状態推定・身体制御・シミュレーション実験を行い, 本手法の有効性を確認する.
  最後に, 本手法と実験結果について考察と結論を述べる.
}%

\section{Musculoskeletal Humanoids and Intersensory Networks} \label{sec:musculoskeletal-introduction}

\subsection{Musculoskeletal Humanoids} \label{subsec:musculoskeletal-humanoids}
\switchlanguage%
{%
  The musculoskeletal humanoid covered in this study has a musculoskeletal structure mimicking human joints and muscles \cite{nakanishi2013design, wittmeier2013toward, jantsch2013anthrob, asano2016kengoro}.
  Compared to the tendon-driven robot with constant moment arms of muscles to joints \cite{kobayashi1998tendon, niiyama2010athlete} which is easy to modelize, these robots are difficult to modelize because the moment arms change complexly depending on the posture.
  We show the basic structure of musculoskeletal humanoids in the upper left figure of \figref{figure:musculoskeletal-humanoid}.
  Muscles are antagonistically arranged around joints, and muscle length $l$, muscle tension $T$, and muscle temperature $C$ of each muscle can be measured.
  Dyneema, which is a chemical fiber resistant to friction, is used for each muscle and it slightly elongates depending on muscle tension.
  Depending on the robot, there are nonlinear elastic elements at the ends of muscles in order to enable variable stiffness control.
  Also, although some robots have sensors to measure joint angles $\theta$ \cite{kawaharazuka2019musashi, urata2006sensor}, ordinary musculoskeletal humanoids do not have these sensors due to complex joint structures such as ball joints and the scapula, and so joint angles cannot be directly measured.
  However, the actual joint angles can be estimated by using vision sensor attached to the head \cite{kawaharazuka2018online}.
  In this method, an AR maker is attached to the robot end effector, e.g. the hand, to obtain the posture $\bm{P}_{marker}$, and Inverse Kinematics is solved to $\bm{P}_{target} = \bm{P}_{marker}$ by setting the initial joint angles $\bm{\theta}_{initial}$ as the estimated joint angles $\bm{\theta}^{estimated}_{current}$ from the change in muscle lengths \cite{ookubo2015learning}, as shown below,
  \begin{align}
    \bm{\theta}^{estimated'}_{current} = \textrm{IK}(\bm{P}_{target}=\bm{P}_{marker}, \bm{\theta}_{initial}=\bm{\theta}^{estimated}_{current})
  \end{align}
  where IK means Inverse Kinematics, and $\bm{\theta}^{estimated'}_{current}$ is the estimated actual joint angles compensated for by vision sensor.
  There are other sensors such as IMUs, tactile sensors, etc.

  We show the musculoskeletal humanoid Musashi \cite{kawaharazuka2019musashi} used for experiments of this study in the center figure of \figref{figure:musculoskeletal-humanoid}, and show the muscle arrangement of its left arm in the right figure of \figref{figure:musculoskeletal-humanoid}.
  Musashi is mainly composed of the muscle modules, joint modules, nonlinear elastic elements shown in the lower left figure of \figref{figure:musculoskeletal-humanoid}.
  The joint modules enabling the measurement of joint angles advance our investigation of learning control systems, and nonlinear elastic elements enable soft contact with the environment and variable stiffness control.
  In this study, we mainly handle the 3 degrees of freedom (DOFs) shoulder and 2 DOFs elbow, and we represent these joint angles as $\bm{\theta}=(\theta_{S-r}, \theta_{S-p}, \theta_{S-y}, \theta_{E-p}, \theta_{E-y})$ ($S$ means the shoulder, $E$ means the elbow, and $rpy$ means the roll, pitch, and yaw, respectively).
  Also, 10 muscles are related to these 5 DOFs.
}%
{%
  本研究で扱う筋骨格ヒューマノイドは, 人体を模倣した関節と筋構造を有する\cite{nakanishi2013design, wittmeier2013toward, jantsch2013anthrob, asano2016kengoro}.
  これは, 関節に対する筋のモーメントアームが一定等のモデル化が容易な腱駆動型ロボット\cite{kobayashi1998tendon, niiyama2010athlete}とは違い, 筋のモーメントアームが複雑に変化しモデル化が困難であるという特徴を持つ.
  筋骨格ヒューマノイドの基本的な構造を\figref{figure:musculoskeletal-humanoid}の左上図に示す.
  関節に対して筋が拮抗に配置されており, それぞれの筋において筋長$l$, 筋張力$T$を測定することができる.
  筋は摩耗に強い化学繊維であるDyneemaを使用しており, 筋自体が多少伸びると同時に, ロボットによっては筋の末端に非線形弾性要素が存在し可変剛性制御が可能となっている.
  また, ロボットによっては関節角度$\theta$を測定することが可能だが\cite{kawaharazuka2019musashi, urata2006sensor}, 通常の筋骨格ヒューマノイドでは球関節や肩甲骨等の複雑な関節構造から, 直接その関節角度を測定することはできない.
  しかし, 視覚センサを用いることで, \cite{kawaharazuka2018online}と同様に実機関節角度を推定することができ, これを実機値として用いることも可能である.
  これは以下のように, ロボットのエンドエフェクタにARマーカ等を取り付け, その位置$\bm{P}_{marker}$を取得し, 筋長変化からの関節角度推定値$\bm{\theta}^{estimated}_{current}$ \cite{ookubo2015learning}を初期値$\bm{\theta}_{initial}$として$\bm{P}_{target}=\bm{P}_{marker}$に対して逆運動学を解くという手法である.
  \begin{align}
    \bm{\theta}^{estimated'}_{current} = \textrm{IK}(\bm{P}_{target}=\bm{P}_{marker}, \bm{\theta}_{initial}=\bm{\theta}^{estimated}_{current})
  \end{align}
  ここで, IKは逆運動学を, $\bm{\theta}^{estimated'}_{current}$は視覚により補正された実機関節角度推定値である.
  その他, 慣性センサや接触センサ等が存在する場合もある.

  本研究の実験において用いる筋骨格ヒューマノイドMusashiを\figref{figure:musculoskeletal-humanoid}に中図に, その左腕に関する筋配置を右図に示す.
  Musashiは, \figref{figure:musculoskeletal-humanoid}の左下図に示す筋モジュール・関節モジュール・非線形弾性要素等から成り立つ.
  関節角度を測定可能な関節モジュールにより学習制御模索を促進し, かつ非線形弾性要素により環境との柔らかな接触や可変剛性制御等を可能としている.
  本研究では主に, 肩の3自由度, 肘の2自由度を用いて実験を行うこととし, 関節角度は$\bm{\theta}=(\theta_{S-r}, \theta_{S-p}, \theta_{S-y}, \theta_{E-p}, \theta_{E-y})$のように表す($S$はshoulder, $E$はelbow, $rpy$はそれぞれroll, pitch, yawを表す).
}%

% \begin{figure}[t]
%   \centering
%   \includegraphics[width=1.0\columnwidth]{figs/intersensory-networks}
%   \caption{The relationship among sensors included in the musculoskeletal humanoid.}
%   \label{figure:intersensory-networks}
%   \vspace{-1.0zh}
% \end{figure}

\subsection{Musculoskeletal Intersensory Networks} \label{subsec:intersensory-networks}
\switchlanguage%
{%
  In this study, among the introduced sensors, we focus on the joint angle $\bm{\theta}$, muscle tension $\bm{T}$, and muscle length $\bm{l}$ which can be definitely measured and have strong correlations.
  In previous studies \cite{mizuuchi2006acquisition, kawaharazuka2018online}, the nonlinear relationship between $\bm{\theta}$ and $\bm{l}$: $\bm{\theta}\to\bm{l}$ is trained.
  By extending them and considering the body tissue softness specific to the musculoskeletal structure, the relationship of $(\bm{\theta}, \bm{T})\to\bm{l}$ is trained in \cite{kawaharazuka2018bodyimage, kawaharazuka2019longtime}.

  In actuality, there exist relationships of not only \textcircled{\scriptsize1} $(\bm{\theta}, \bm{T})\to\bm{l}$ but also \textcircled{\scriptsize2} $(\bm{T}, \bm{l})\to\bm{\theta}$ and \textcircled{\scriptsize3} $(\bm{\theta}, \bm{l})\to\bm{T}$.
  Thus, among $(\bm{\theta}, \bm{T}, \bm{l})$, one sensor value can be calculated from the other two.
  In other words, all of the sensor information can be calculated from two sensor values.

  Of course, other sensors such as vision and tactile information will be able to be incorporated in the future.
}%
{%
  本研究では, これまで述べたセンサの中でも基本的に測定可能な関節角度$\bm{\theta}$, 筋張力$\bm{T}$, 筋長$\bm{l}$に焦点を当てる.
  これまで, \cite{mizuuchi2006acquisition, kawaharazuka2018online}においては$\bm{\theta}$と$\bm{l}$の非線形な関係$\bm{\theta}\to\bm{l}$を学習していた.
  さらにそれを発展させ, \cite{kawaharazuka2018bodyimage}では筋骨格構造に特有な身体組織の柔軟性を考慮し, $(\bm{\theta}, \bm{T})\to\bm{l}$を学習していた.

  実際には, \figref{figure:intersensory-networks}に示すように\textcircled{\scriptsize1} $(\bm{\theta}, \bm{T})\to\bm{l}$だけでなく, \textcircled{\scriptsize2} $(\bm{T}, \bm{l})\to\bm{\theta}$や\textcircled{\scriptsize3} $(\bm{l}, \bm{\theta})\to\bm{T}$という関係が存在する.
  つまり, $(\bm{\theta}, \bm{T}, \bm{l})$の3つのセンサ情報のうち, 2つから残り1つを求めることができるということである.
  これは言い換えれば, $(\bm{\theta}, \bm{T}, \bm{l})$のうち2つのセンサ情報から, 3つ全てのセンサ情報を取得できることに相当する.
}%

\begin{figure}[t]
  \centering
  \includegraphics[width=0.8\columnwidth]{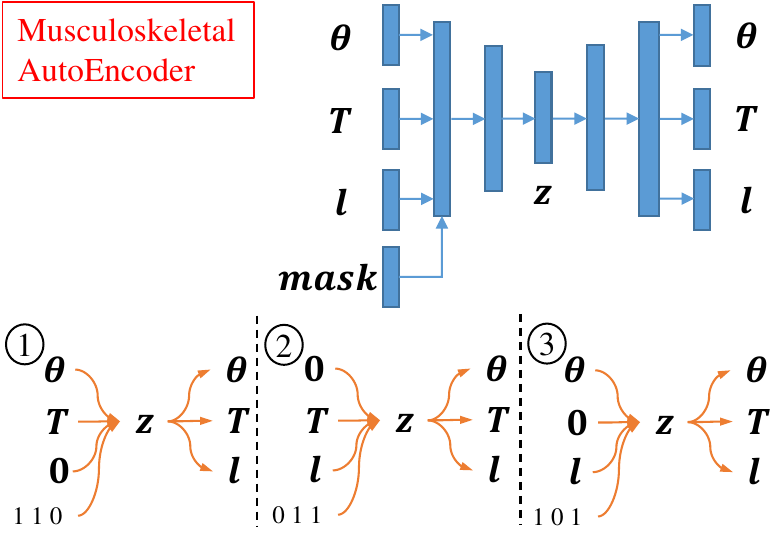}
  \caption{The network structure of Musculoskeletal AutoEncoder.}
  \label{figure:musculoskeletal-autoencoder}
  \vspace{-3.0ex}
\end{figure}

\begin{figure*}[t]
  \centering
  \includegraphics[width=1.7\columnwidth]{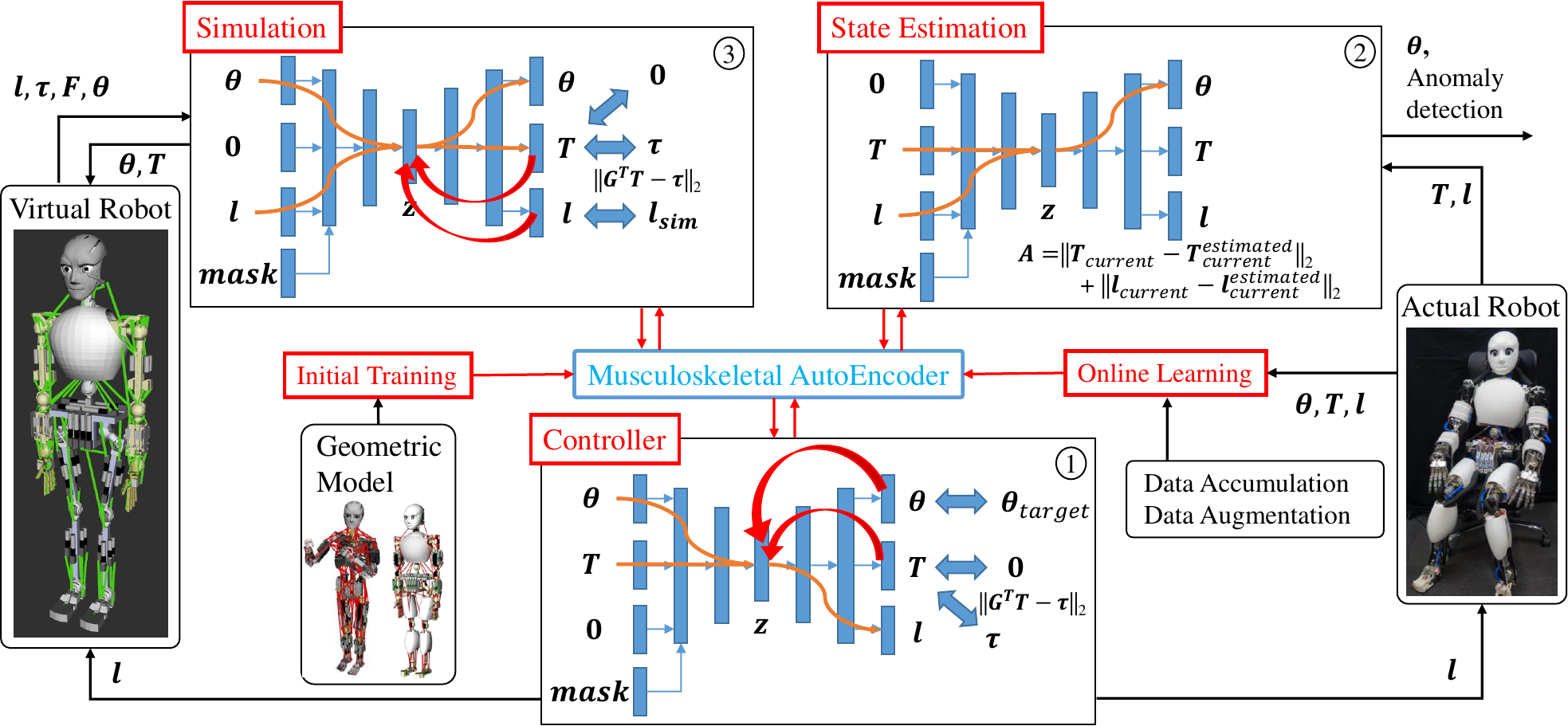}
  \caption{State estimation, control, and simulation methods using Musculoskeletal AutoEncoder.}
  \label{figure:whole-system}
  \vspace{-3.0ex}
\end{figure*}

\section{Musculoskeletal AutoEncoder} \label{sec:musculoskeletal-autoencoder}
\subsection{Network Structure and System Overview} \label{subsec:network-structure}
\switchlanguage%
{%
  First, we show the network structure of MAE in \figref{figure:musculoskeletal-autoencoder}.
  MAE is a simple AutoEncoder-type network \cite{hinton2006reducing} whose inputs are $(\bm{\theta}, \bm{T}, \bm{l})$ and $\bm{mask}$, and whose output is $(\bm{\theta}, \bm{T}, \bm{l})$.
  The dimension of $(\bm{\theta}, \bm{T}, \bm{l}, \bm{mask})$ is $(D, M, M, 3)$ ($D$ is the number of joint angles, $M$ is the number of muscles, and $\bm{mask}$ has 3 binarized values).
  We represent the middle layer of MAE, whose dimension is the lowest, as $z$, and its dimension is $D+M$.
  The reason is that the dimension of $z$ can be reduced to less than $D+M$ because $(\bm{\theta}, \bm{T}, \bm{l})$ can be inferred from $(\bm{\theta}, \bm{T})$, $D+M$ dimension.
  In this study, MAE is composed of 7 layers including its input and output layers, and the numbers of units are $(D+2M+3, 200, 30, D+M, 30, 200, D+2M)$.
  The number of layers do not largely affect the final inference accuracy, but the inference accuracy decreases with less number of units in the hidden layers than this configuration.
  Also, in order to align the scales of $(\bm{\theta}, \bm{T}, \bm{l})$, we set their units as ([rad], [N/200], [mm/100]).

  Second, we will explain the overview of initial training and online learning of MAE.
  Although the structure of MAE is similar to an ordinary AutoEncoder, its training method is specific.
  As stated in \secref{subsec:intersensory-networks}, we make use of the characteristic that all 3 sensor values $(\bm{\theta}, \bm{T}, \bm{l})$ can be obtained from two among them.
  Thus, in the case only $(\bm{\theta}, \bm{T})$ can be obtained, like \textcircled{\scriptsize1} shown in \figref{figure:musculoskeletal-autoencoder}, MAE outputs $(\bm{\theta}, \bm{T}, \bm{l})$ from the input of $(\bm{\theta}, \bm{T}, \bm{0}, (1, 1, 0))$.
  In the beginning, we conduct initial training of MAE from the geometric model including the relationship between joint angles and muscle lengths.
  Next, we update MAE online using the actual robot sensor information.

  We represent the encoder part of MAE as $\bm{z}=\bm{f}_{encode}(\bm{\theta}, \bm{T}, \bm{l}, \bm{mask})$ and the decoder part of MAE as $(\bm{\theta}, \bm{T}, \bm{l}) = \bm{f}_{decode}(\bm{z})$.
  Also, we represent $\bm{f}_{encode}$ of \textcircled{\scriptsize1} as $\bm{f}_{encode, 1}(\bm{\theta}, \bm{T})=\bm{f}_{encode}(\bm{\theta}, \bm{T}, \bm{0}, (1, 1, 0))$, that of \textcircled{\scriptsize2} as $\bm{f}_{encode, 2}(\bm{T}, \bm{l})=\bm{f}_{encode}(\bm{0}, \bm{T}, \bm{l}, (0, 1, 1))$, and that of \textcircled{\scriptsize3} as $\bm{f}_{encode, 3}(\bm{\theta}, \bm{l})=\bm{f}_{encode}(\bm{\theta}, \bm{0}, \bm{l}, (1, 0, 1))$.
  Regarding the decoder, we represent $\bm{\theta} = \bm{f}_{decode, \bm{\theta}}(\bm{z})$, $\bm{T} = \bm{f}_{decode, \bm{T}}(\bm{z})$, and $\bm{l} = \bm{f}_{decode, \bm{l}}(\bm{z})$.

  We show the whole system using this MAE in \figref{figure:whole-system}.
  By using MAE and its backpropagation, we can conduct state estimation, control, and simulation of musculoskeletal humanoids.
}%
{%
  \figref{figure:musculoskeletal-autoencoder}にMusculoskeletal AutoEncoderのネットワーク構造示す.
  その構造はシンプルであり, $(\bm{\theta}, \bm{T}, \bm{l})$と$\bm{mask}$を入力とし, 同様の$(\bm{\theta}, \bm{T}, \bm{l})$を出力するような, AutoEncoder型\cite{hinton2006reducing}のネットワークとなっている.
  ここで, 関節数を$D$, 筋数を$M$とすると, $(\bm{\theta}, \bm{T}, \bm{l}, \bm{mask})$の次元数はそれぞれ$(D, M, M, 3)$となる.
  Musculoskeletal AutoEncoderにおいて最もユニット数の少ない中間層の値を$z$とし, この次元は$D+M$とする.
  これは, $(\bm{\theta}, \bm{T})$から$(\bm{\theta}, \bm{T}, \bm{l})$の全ての値を推論できるため, 最小のユニット数は$D+M$よりも小さくすることが可能であるためである.
  本研究ではMusculoskeletal AutoEncoderは入力と出力層を含めた7層とし, そのユニット数は$(D+2M+3, 200, 30, D+M, 30, 200, D+2M)$としている.
  本研究では層の数はそこまで大きく最終的な結果には影響しなかったが, ユニット数に関しては, これ以上減らすと, 精度が落ちていった.
  また, スケールを揃えるため$\bm{\theta}$の単位は[rad], $\bm{T}$の単位は[N/200], $\bm{l}$の単位は[mm/100]としている.

  次に, Musculoskeletal AutoEncoderの初期学習・オンライン学習の概要について述べる.
  Musculoskeletal AutoEncoderは通常のAutoEncoderと構造は似ているものの, 学習方法は特殊である.
  \secref{subsec:intersensory-networks}で述べたように, $(\bm{\theta}, \bm{T}, \bm{l})$のうち2つのセンサ情報から3つのセンサ情報全てを取得できるという性質を利用する.
  つまり, \figref{figure:musculoskeletal-autoencoder}の\textcircled{\scriptsize1}のように, 得られるセンサ情報が$(\bm{\theta}, \bm{T})$のみの場合は$(\bm{\theta}, \bm{T}, \bm{0}, (1, 1, 0))$を入力として, $(\bm{\theta}, \bm{T}, \bm{l})$を出力する.
  \textcircled{\scriptsize2}, \textcircled{\scriptsize3}も同様である.
  まず関節と筋の関係を含む幾何モデルをからMusculoskeletal AutoEncoderを初期学習させる.
  その後, 実機センサデータを取得し, このネットワークをオンラインで更新していくこととなる.

  Musculoskeletal AutoEncoderにおけるEncode部分を$\bm{z}=\bm{f}_{encode}(\bm{\theta}, \bm{T}, \bm{l}, \bm{mask})$, Decode部分を$(\bm{\theta}, \bm{T}, \bm{l}) = \bm{f}_{decode}(\bm{z})$とする.
  また, \textcircled{\scriptsize1}における$\bm{f}_{encode}$を$\bm{f}_{encode, 1}(\bm{\theta}, \bm{T})=\bm{f}_{encode}(\bm{\theta}, \bm{T}, \bm{0}, (1, 1, 0))$とし, その他同様に\textcircled{\scriptsize2}については$\bm{f}_{encode, 2}(\bm{T}, \bm{l})$, \textcircled{\scriptsize3}については$\bm{f}_{encode, 3}(\bm{\theta}, \bm{l})$とする.
  Decode部分に関しても, $\bm{\theta} = \bm{f}_{decode, \bm{\theta}}(\bm{z})$, $\bm{T} = \bm{f}_{decode, \bm{T}}(\bm{z})$, $\bm{l} = \bm{f}_{decode, \bm{l}}(\bm{z})$とする.

  \figref{figure:whole-system}にMusculoskeletal AutoEncoderを用いた全体システム図を示す.
  Musculoskeletal AutoEncoderと誤差逆伝播をうまく用いることで, 筋骨格ヒューマノイドの状態推定・制御・シミュレーションを行うことができる.
}%

\subsection{Initial Training} \label{subsec:initial-training}
\switchlanguage%
{%
  We will explain the initial training of MAE using a geometric model.
  In the geometric model, each muscle route is represented by linearly connecting its start point, relay points, and end point.
  By setting a certain joint angle, muscle length can be calculated from the distance of the start, relay, and end points.
  In this study, we represent a function calculating the absolute muscle length from the geometric model as $\bm{f}_{geo, abs}(\bm{\theta})$, and represent a function calculating the muscle length relative to the length at the initial posture (all the joint angles are 0, $\bm{\theta}=\bm{0}$) as $\bm{f}_{geo}(\bm{\theta})$.

  First, we estimate the relationship between muscle tension $\bm{T}$ and muscle length elongation including nonlinear elastic elements $\bm{L}_{e}(\bm{l}_{abs}, \bm{T})$ considering the elongation of Dyneema, which is proportional to the absolute muscle length $\bm{l}_{abs}$, from a testbed of one muscle.
  Second, we generate the data as shown below,
  \begin{align}
    (\bm{\theta}, \bm{T}, \bm{l}) = (\bm{\theta}_{r}, \bm{T}_{r}, \bm{f}_{geo}(\bm{\theta}_{r})-\bm{L}_{e}(\bm{f}_{geo, abs}(\bm{\theta}_{r}), \bm{T}_{r}))
  \end{align}
  where $\bm{\theta}_{r}$ is random joint angle, and $\bm{T}_{r}$ is random muscle tension.
  Using this data, we randomly execute \textcircled{\scriptsize1}, \textcircled{\scriptsize2}, and \textcircled{\scriptsize3} shown in \figref{figure:musculoskeletal-autoencoder}.
  Thus, we randomly set one element among $(\bm{\theta}, \bm{T}, \bm{l})$ as $\bm{0}$, input the corresponding $\bm{mask}$ into MAE, and output $(\bm{\theta}_{calc}, \bm{T}_{calc}, \bm{l}_{calc})$.
  Finally, we calculate the loss $L_{initial}$ as shown below, and train MAE,
  \begin{align}
    L_{initial} = w_{\theta}||\bm{\theta}-\bm{\theta}_{calc}||_{2} + w_{T}||\bm{T}-\bm{T}_{calc}||_{2} + w_{l}||\bm{l}-\bm{l}_{calc}||_{2}
  \end{align}
  where $||\cdot||_{2}$ represents L2 norm.

  In this study, we set the number of batches for initial training $C^{initial}_{batch}=100$, the number of epochs $C^{initial}_{epoch}=30$, $w_{\theta}=1.0$, $w_{T}=10.0$, and $w_{l}=10.0$.
  Also, we use Adam \cite{kingma2015adam} as the optimization method of MAE.
}%
{%
  幾何モデルを用いたMusculoskeletal AutoEncoderの初期学習について述べる.
  ここで言う幾何モデルとは, 筋の起始点・中継点・終止点を直線で結び筋経路を表現したものである.
  関節角度を指定すると, 筋の経由点間の距離から筋長を求めることができる.
  ここで, 幾何モデルから筋の絶対長さを求める関数を$\bm{f}_{geo, abs}(\bm{\theta})$, 初期姿勢(全関節角度が0, つまり$\bm{\theta}=\bm{0}$)からの相対筋長を求める関数を$\bm{f}_{geo}(\bm{\theta})$とする.

  まず, 筋単体のテストベッドにより, 筋長の全体長さ$\bm{l}_{abs}$に比例するダイニーマの伸びを考慮した, 筋張力$\bm{T}$と非線形弾性要素を含む筋長変化の関係$\bm{L}_{e}(\bm{l}_{abs}, \bm{T})$を関数フィッティングにより求める.
  次に, 以下のようなデータを生成する.
  \begin{align}
    (\bm{\theta}, \bm{T}, \bm{l}) = (\bm{\theta}_{r}, \bm{T}_{r}, \bm{f}_{geo}(\bm{\theta}_{r})-\bm{L}_{e}(\bm{f}_{geo, abs}(\bm{\theta}_{r}), \bm{T}_{r}))
  \end{align}
  ここで, $\bm{\theta}_{r}$はランダムな関節角度, $\bm{T}_{r}$はランダムな筋張力である.
  このデータを用いて, \figref{figure:musculoskeletal-autoencoder}における\textcircled{\scriptsize1}\textcircled{\scriptsize2}\textcircled{\scriptsize3}をランダムに実行する.
  つまり, $(\bm{\theta}, \bm{T}, \bm{l})$のうち一つをランダムに$\bm{0}$にし, 対応する$\bm{mask}$を加えたデータをネットワークの入力とし, $(\bm{\theta}_{calc}, \bm{T}_{calc}, \bm{l}_{calc})$を出力する.
  最後に, 損失関数$L_{initial}$を以下のように計算し, Musculoskeletal AutoEncoderを学習させる.
  \begin{align}
    L_{initial} = w_{\theta}||\bm{\theta}-\bm{\theta}_{calc}||_{2} + w_{T}||\bm{T}-\bm{T}_{calc}||_{2} + w_{l}||\bm{l}-\bm{l}_{calc}||_{2}
  \end{align}
  ここで, $||\cdot||_{2}$はL2ノルムを表す.

  本研究では学習の際のバッチ数を$C^{initial}_{batch}=100$, エポック数を$C^{initial}_{epoch}=30$, $w_{\theta}=1.0$, $w_{T}=10.0$, $w_{l}=10.0$とし, 最適化手法はAdam \cite{kingma2015adam}を用いる.

  % 先行研究である$(\bm{\theta}, \bm{T})\to\bm{l}$を求める手法\cite{kawaharazuka2018bodyimage}では, そのスケールの大きさの違いから, 外力の働かない理想的な状態における関節と筋の関係を表すネットワーク$\bm{f}_{ideal}$と, 身体組織の柔軟性による筋経路変化補償項$\bm{g}(\bm{\theta}, \bm{T})$のネットワークを分けて別々に学習していた.
  % 本研究では, これらを一つにする必要があり, 初期学習は二段階で行われる.

  % まず, $\bm{g}$と同じ筋経路変化補償項を表すネットワークの構造を形成させる.
  % 筋単体のテストベッドにより, 筋長の全体長さ$\bm{l}_{abs}$に比例するダイニーマの伸びを考慮した, 筋張力$\bm{T}$と非線形弾性要素を含む筋長変化の関係$\bm{L}_{e}(\bm{l}_{abs}, \bm{T})$を関数フィッティングにより求める.
  % その後, 以下のようなデータを用いてMusculoskeletal AutoEncoderを学習させる.
  % \begin{align}
  %   (\bm{\theta}, &\bm{T}, \bm{l}, \bm{mask})\nonumber\\
  %   &= (\bm{\theta}_{r}, \bm{T}_{r}, -\bm{L}_{e}(\bm{f}_{geo, abs}(\bm{\theta}_{r}), \bm{T}_{r}), \bm{mask}_{1})
  % \end{align}
  % ここで, $\bm{\theta}_{r}$はランダムな関節角度, $\bm{T}_{r}$はランダムな筋張力, $\bm{mask}_{1}$は$(1, 1, 0)$であり, これは\figref{figure:musculoskeletal-autoencoder}の\textcircled{\scriptsize1}に相当する.

  % 次に, $\bm{f}_{ideal}$と$\bm{g}$と合わせ, 以下のようなデータを用いて全体を学習させる.
  % \begin{align}
  %   (\bm{\theta}, &\bm{T}, \bm{l}, \bm{mask})\nonumber\\
  %   &= (\bm{\theta}_{r}, \bm{T}_{r}, \bm{f}_{geo}(\bm{\theta}_{r})-\bm{L}_{e}(\bm{f}_{geo, abs}(\bm{\theta}_{r}), \bm{T}_{r}), \bm{mask}_{r})
  % \end{align}
  % ここで, $\bm{mask}_{r}$は$(1, 1, 0), (0, 1, 1), (1, 0, 1)$の3つからランダムに選んだ値である.
  % これらの段階が必要なのは, $\bm{f}_{ideal}+\bm{g}$を最初から同時に学習してしまうと, 値のスケールの小さな$\bm{g}$の影響を上手く学習することが難しいためである.

  % 本研究では第一段階におけるバッチ数を$C^{1}_{batch}=100$, エポック数を$C^{1}_{epoch}=30$とし, 第二段階におけるバッチ数を$C^{2}_{batch}=100$, エポック数を$C^{2}_{epoch}=100$とする.
  % また, 最適化手法はAdam \cite{kingma2015adam}を用いる.
}%

\subsection{Online Learning} \label{subsec:online-learning}
\switchlanguage%
{%
  We will explain the online learning method of MAE using the actual robot sensor information.

  First, the sensor values of $(\bm{\theta}_{current}, \bm{T}_{current}, \bm{l}_{current})$ are accumulated from the actual robot.
  If joint angles cannot be directly measured, $\bm{\theta}^{estimated'}_{current}$ in \secref{subsec:musculoskeletal-humanoids} is used as $\bm{\theta}_{current}$.
  The data is accumulated when the current joint angle or muscle tension deviates from the previously accumulated data by more than a certain threshold, and the robot is not moving.
  MAE begins to be updated when the number of the accumulated data exceeds $C_{thre}$.
  A batch for online learning is constructed using the newest data, $(\bm{0}, \bm{0}, \bm{0})$, and randomly chosen $C_{data}$ ($C_{data}\leq{C}_{thre}$) number of data from the accumulated data.
  Then, 3 kinds of $\bm{mask}$ of \textcircled{\scriptsize1}, \textcircled{\scriptsize2}, and \textcircled{\scriptsize3} are added to each data in the batch.
  Finally, MAE is updated using the batch with $3\times(C_{data}+2)$ number of data in total.

  In this study, we set the number of batches for online learning $C^{online}_{batch}=10$, the number of epochs $C^{online}_{epoch}=10$, $C_{thre}=20$, and $C_{data}=10$.
}%
{%
  実機センサデータを用いたオンライン学習について述べる.

  まず, 実機からセンサデータ$(\bm{\theta}_{current}, \bm{T}_{current}, \bm{l}_{current})$を取得する.
  ただし, これは関節角度か筋張力が前回の学習の際のデータからある閾値以上離れ, かつ静止している場合に学習データを作成する.

  次に, このデータを蓄積・拡張し, musculoskeletal autoencoderをオンラインで更新していく.
  データ$(\bm{\theta}_{current}, \bm{T}_{current}, \bm{l}_{current})$を蓄積していき, データ数が$C_{thre}$を超えたところで, 更新を開始する.
  蓄積したデータの中から$C_{data}$ ($C_{data}\leq{C}_{thre}$)個, そして最新のデータ1個を取得する.
  また, $(\bm{0}, \bm{0}, \bm{0})$というデータも一個加える.
  % 最後に, $\bm{t}$による筋組織の変化項は同じ$\bm{\theta}$周辺では大きくは変化しないという仮定から, これまで取り出した$c_{data}+2$個のデータ$(\bm{\theta}, \bm{t}, \bm{l})$を加工した$(\bm{\theta}_{around}, \bm{t}, \bm{l}-\bm{f}_{geo}(\bm{\theta})+\bm{f}_{geo}(\bm{\theta}_{around}))$も使用する.
  % ここで, $\bm{\theta}_{around}$とは, $\bm{\theta}$に対して平均0, 分散$c_{var}$の正規分布に従う乱数を加えた値とする.
  これらデータに対して, それぞれ\textcircled{\scriptsize1}\textcircled{\scriptsize2}\textcircled{\scriptsize3}の3種類の$\bm{mask}$の値を加え, 計$3\times(C_{data}+2)$個のデータを用いてネットワークを更新する.

  本研究では, 更新の際のバッチ数を$C^{online}_{batch}=10$, エポック数を$C^{online}_{epoch}=10$とし, その他定数は$C_{thre}=20$, $C_{data}=10$とした.
}%

\subsection{State Estimation Using Musculoskeletal AutoEncoder} \label{subsec:estimation-method}
\switchlanguage%
{%
  We will explain the state estimation method of musculoskeletal humanoids using the trained MAE.
  As stated in \secref{subsec:musculoskeletal-humanoids}, the ordinary musculoskeletal humanoid \cite{asano2016kengoro} does not have joint angle sensors, and so the actual joint angles are estimated using vision information.
  However, in this case, an AR marker has to be attached to the robot, and the robot must keep watching its body to know the current joint angles.
  By estimating joint angles from the current muscle length and muscle tension information, the robot can know the self-body state continuously.

  The joint angle estimation method using MAE is much simpler than the previous method \cite{ookubo2015learning}.
  As shown in the upper right figure of \figref{figure:whole-system}, the current muscle tension $\bm{T}_{current}$ and muscle length $\bm{l}_{current}$ are fed into MAE like \textcircled{\scriptsize2} shown in \figref{figure:musculoskeletal-autoencoder}, and the estimated joint angles $\bm{\theta}^{estimated}_{current} = \bm{f}_{decode, \bm{\theta}}(\bm{f}_{encode, 2}(\bm{T}_{current}, \bm{l}_{current}))$ can be obtained.

  Also, although the previous methods can only conduct joint angle estimation, MAE can be used for anomaly detection, too.
  When estimating joint angles, not only $\bm{\theta}^{estimated}_{current}$ but also all of the values of $(\bm{\theta}^{estimated}_{current}, \bm{T}^{estimated}_{current}, \bm{l}^{estimated}_{current}) = \bm{f}_{decode}(\bm{f}_{encode, 2}(\bm{T}_{current}, \bm{l}_{current}))$ can be obtained.
  In this procedure, the ouput of MAE is trained to restore the input value, and so $(\bm{T}^{estimated}_{current}, \bm{l}^{estimated}_{current})$ should be the same value as $(\bm{T}_{current}, \bm{l}_{current})$.
  In an anomaly case such that muscles rupture, restoring the input value is difficult because the data in this case is not used to train MAE.
  Therefore, the value $A$, stated as below, rises.
  \begin{align}
    A = ||\bm{T}^{estimated}_{current}-\bm{T}_{current}||_{2} + ||\bm{l}^{estimated}_{current}-\bm{l}_{current}||_{2}
  \end{align}
  By monitoring $A$, we can detect anomaly, stop online learning, and stop movements.
}%
{%
  まず状態推定の手法について述べる.
  筋骨格ヒューマノイドは\secref{subsec:musculoskeletal-humanoids}で述べたように一般的には関節角度を測定することができないため, 視覚を用いて実機関節角度を推定する.
  しかしこの場合, 身体にマーカ等をつける必要や, 身体を常に見続けなければならない等の制約を受けることになる.
  そこで, 現在の筋張力・筋長情報から, 関節角度を推定することで, 自身の状態を常に知り続けることが可能となる.

  Musculoskeletal AutoEncoderを用いた関節角度推定は\cite{ookubo2015learning}等に比べて非常に単純である.
  \figref{figure:whole-system}の左図に示すように, 実機から現在筋張力$\bm{T}_{current}$, 現在筋長$\bm{l}_{current}$を取得し, これを\figref{figure:musculoskeletal-autoencoder}の\textcircled{\scriptsize2}の形でネットワークに入力し, 関節角度推定値$\bm{\theta}^{estimated}_{current} = \bm{f}_{decode, \bm{\theta}}(\bm{f}_{encode, 2}(\bm{T}_{current}, \bm{l}_{current}))$を得るのみである.

  また, これまでの手法では関節角度推定のみしか出来なかったが, Musculoskeletal AutoEncoderを用いることで異常検知をすることができる.
  先ほどの関節角度推定の際, $\bm{\theta}^{estimated}_{current}$だけでなく, $(\bm{\theta}^{estimated}_{current}, \bm{T}^{estimated}_{current}, \bm{l}^{estimated}_{current}) = \bm{f}_{decode}(\bm{f}_{encode, 2}(\bm{T}_{current}, \bm{l}_{current}))$のように, 全ての値を取り出す.
  ここで, Musculoskeletal AutoEncoderの\textcircled{\scriptsize2}では$\bm{T}, \bm{l}$は入力の値と同じ値を復元するように学習されている.
  しかし, 例えば筋が切れた等の異常が発生した場合は, その状態を学習していないため復元が難しくなる.
  よって, 以下の値が上昇する.
  \begin{align}
    A = ||\bm{T}^{estimated}_{current}-\bm{T}_{current}||_{2} + ||\bm{l}^{estimated}_{current}-\bm{l}_{current}||_{2}
  \end{align}
  この値$A$を監視することによって, オンライン学習を停止したり, 異常を検知して動作を停止することが可能となる.
}%

\subsection{Control Using Musculoskeletal AutoEncoder} \label{subsec:control-method}
\switchlanguage%
{%
  We will explain two muscle length-based controls using MAE: the same one as \cite{kawaharazuka2018bodyimage} and a novel one with backpropagation, and then explain the overview of muscle tension-based control and variable stiffness control.
  We mainly use the novel muscle length-based control with backpropagation in our experiments.

  First, we will explain a method to conduct the same control as \cite{kawaharazuka2018bodyimage} using MAE.
  As an assumption, musculoskeletal humanoids are moved using muscle stiffness control (MSC) \cite{shirai2011stiffness} as shown below,
  \begin{align}
    \bm{T}_{target} = \bm{T}_{bias} + \textrm{max}(\bm{0}, K_{stiff}(\bm{l}_{current}-\bm{l}_{target})) \label{eq:muscle-stiffness-control}
  \end{align}
  where $\bm{T}_{target}$ is target muscle tension, $\bm{T}_{bias}$ is a bias value of MSC, and $K_{stiff}$ is a muscle stiffness gain.
  The larger $K_{stiff}$ is, the less the error between the target and actual muscle lengths can be permitted, and the smaller $K_{stiff}$ is, the worse the tracking ability becomes.
  In order to realize a certain target joint angle $\bm{\theta}_{target}$, $\bm{l}_{target}$ of \equref{eq:muscle-stiffness-control} needs to be decided as shown below,
  \begin{align}
    &\bm{l}_{comp}(\bm{T}) = -(\bm{T} - \bm{T}_{bias})/K_{stiff} \\
    &\bm{l}_{target} = \bm{f}_{dec, \bm{l}}(\bm{f}_{enc, 1}(\bm{\theta}_{target}, \bm{T}_{const}))+\bm{l}_{comp}(\bm{T}_{const}) \label{eq:move-first}\\
    &\bm{l}_{target} = \bm{f}_{dec, \bm{l}}(\bm{f}_{enc, 1}(\bm{\theta}_{target}, \bm{T}_{current}))+\bm{l}_{comp}(\bm{T}_{current}) \label{eq:move-second}
  \end{align}
  where $\bm{T}_{const}$ is a certain constant muscle tension (the same value as $\bm{T}_{bias}$ in this study), $\bm{l}_{comp}$ is a compensation value of muscle elongation from $\bm{l}_{target}$ by MSC, and $\bm{f}_{enc}$ and $\bm{f}_{dec}$ are the abbreviation of $\bm{f}_{encode}$ and $\bm{f}_{decode}$, respectively.
  As in \cite{kawaharazuka2018bodyimage}, by conducting \equref{eq:move-first} and \equref{eq:move-second} in order, the robot can realize $\bm{\theta}_{target}$.

  Second, we will explain a novel way to control musculoskeletal humanoids using MAE.
  We first obtain the state of latent space $\bm{z}=\bm{f}_{encode, 1}(\bm{\theta}_{target}, \bm{T}_{current})$ from $\bm{T}_{current}$ and $\bm{\theta}_{target}$.
  Next, the procedures shown below are repeated,
  \begin{enumerate}
    \item Calculate $(\bm{\theta}_{calc}, \bm{T}_{calc}, \bm{l}_{calc}) = \bm{f}_{dec}(\bm{z})$.
    \item Calculate the loss of $L_{control}(\bm{\theta}_{calc}, \bm{T}_{calc}, \bm{l}_{calc})$.
    \item Update $\bm{z}$ by backpropagation \cite{rumelhart1986backprop}.
  \end{enumerate}
  In 2), the loss of $L_{control}$ is calculated as shown below,
  \begin{align}
    L_{control}(\bm{\theta}, \bm{T}, \bm{l}) = &w_{1}||\bm{T}||_{2} + w_{2}||\bm{\theta}-\bm{\theta}_{target}||_{2}\nonumber\\ +& w_{3}||\bm{\tau}_{target}+G^{T}(\bm{\theta}_{target}, \bm{T}_{current})\bm{T}||_{2} \label{eq:keep-control-loss}
  \end{align}
  where $\bm{\tau}_{target}$ is target joint torque, which is necessary to keep $\bm{\theta}_{target}$, calculated from the geometric model.
  Also, $G(\bm{\theta}, \bm{T})$ is a muscle Jacobian calculated from the difference of the outputs, when inputting $(\bm{\theta}, \bm{T})$ and $(\bm{\theta}+\Delta\bm{\theta}$, $\bm{T})$, which adds a slight displacement of $\Delta\bm{\theta}$, into MAE by the form of \textcircled{\scriptsize1}.
  In 3), $\bm{z}$ is updated by backpropagation \cite{rumelhart1986backprop} and gradient descent as shown below,
  \begin{align}
    \bm{g} &= dL_{control}/d\bm{z}\\
    \bm{z} &= \bm{z} - \gamma\bm{g}/||\bm{g}||_{2}
  \end{align}
  where $\bm{g}$ is a gradient of $L_{control}$ regarding $\bm{z}$.
  In this procedure, some batches using various $\gamma$ are generated, and the best $\gamma$ at each step is chosen for better optimization.
  We decide a maximum value of $\gamma$ as $\gamma^{control}_{max}$, divide [0, $\gamma^{control}_{max}$] equally into $N^{control}_{batch}$ parts, and generate $N^{control}_{batch}$ number of $\bm{z}$ updated by these $\gamma$.
  1) and 2) are executed again, and $\bm{z}$ with the lowest $L_{control}$ is used.
  These procedures are repeated $C^{control}_{epoch}$ times.
  Then $(\bm{\theta}_{calc}, \bm{T}_{calc}, \bm{l}_{calc}) = \bm{f}_{dec}(\bm{z})$ is calculated.
  Finally $\bm{l}_{target}$ is calculated from the obtained $\bm{l}_{calc}$, $\bm{T}_{calc}$ as shown below, and is sent to the actual robot.
  \begin{align}
    \bm{l}_{target} = \bm{l}_{calc}+\bm{l}_{comp}(\bm{T}_{calc}) \label{eq:move-keep}
  \end{align}
  By using this method, $\bm{l}_{target}$ with minimized muscle tension can be obtained, while realizing $\bm{\theta}_{target}$.
  Although we can directly calculate $\bm{T}_{target} = \bm{T}_{calc}$ by quadratic programming minimizing $||\bm{T}_{target}||_{2}$ and fulfilling $\tau_{target}=-G^T\bm{T}_{target}$, there is no guarantee that $\bm{T}_{target}$ can be realized at the posture of $\bm{\theta}_{target}$, and so we search $\bm{T}_{target}$ and $\bm{l}_{target}$ in the acquired latent space of $\bm{z}$.

  Third, we will explain the method of muscle tension-based control and variable stiffness control.
  Because only muscle Jacobian is required for muscle tension-based controls, we obtain $G$ as stated above, and can conduct these controls as in \cite{kawamura2016jointspace}.
  Also, we can conduct variable stiffness control as in \cite{kawaharazuka2019longtime} by using $\bm{f}_{encode, 1}(\bm{\theta}, \bm{T})$.

  In this study, we set $w_{1}=1.0$, $w_{2}=1.0$, $w_{3}=0.01$, $\gamma^{control}_{max}=0.5$, $C^{control}_{batch}=10$, and $C^{control}_{epoch}=10$.
}%
{%
  ここでは主に2つの筋長指令による位置制御(\cite{kawaharazuka2018bodyimage}とMAEを用いて再現したもの, MAEを使った新しい手法)について述べ, 最後に筋張力制御・可変剛性制御について概要を述べる.

  まず, \cite{kawaharazuka2018bodyimage}と同様の位置制御をMusculoskeletal AutoEncoderを用いて行う方法について説明する.
  前提として, 本研究では筋骨格ヒューマノイドを以下のような筋剛性制御\cite{shirai2011stiffness}により動作させる.
  \begin{align}
    \bm{T}_{target} = \bm{T}_{bias} + \textrm{max}(\bm{0}, K_{stiff}(\bm{l}_{current}-\bm{l}_{target})) \label{eq:muscle-stiffness-control}
  \end{align}
  ここで, $\bm{T}_{target}$は指令筋張力, $\bm{T}_{bias}$は筋剛性制御のバイアス項, $K_{stiff}$は筋剛性制御の筋剛性係数である.
  筋剛性の値は本研究ではある一定値としているが, 大きくし過ぎると筋長が現在値と指令値の誤差を許容できなくなってしまい, 低すぎると追従性が悪くなる.
  ある関節角度$\bm{\theta}_{target}$を実現したい場合, \equref{eq:muscle-stiffness-control}における$\bm{l}_{target}$を決める必要があり, それは以下のように行う.
  \begin{align}
    &\bm{l}_{comp}(\bm{T}) = -(\bm{T} - \bm{T}_{bias})/K_{stiff} \\
    &\bm{l}_{target} = \bm{f}_{dec, \bm{l}}(\bm{f}_{enc, 1}(\bm{\theta}_{target}, \bm{T}_{const}))+\bm{l}_{comp}(\bm{T}_{const}) \label{eq:move-first}\\
    &\bm{l}_{target} = \bm{f}_{dec, \bm{l}}(\bm{f}_{enc, 1}(\bm{\theta}_{target}, \bm{T}_{current}))+\bm{l}_{comp}(\bm{T}_{current}) \label{eq:move-second}
  \end{align}
  ここで, $\bm{T}_{const}$はある一定の筋張力(基本的には$\bm{T}_{bias}$を用いる), $\bm{l}_{comp}$は筋剛性制御によるソフトウェアでの$\bm{l}_{target}$からの筋の伸びを補償するための項, $\bm{f}_{enc}$, $\bm{f}_{dec}$はそれぞれ$\bm{f}_{encode}$, $\bm{f}_{decode}$の省略形である.
  \cite{kawaharazuka2018bodyimage}にあるように, \equref{eq:move-first}, \equref{eq:move-second}の順に実行することで, $\bm{\theta}_{target}$を実現する.

  次に, Musculoskeletal AutoEncoderを用いたより賢い制御手法について説明する.
  まず, 現在の筋張力$\bm{T}_{current}$, 指令関節角度$\bm{\theta}_{target}$から, 潜在状態$\bm{z}=\bm{f}_{encode, 1}(\bm{\theta}_{target}, \bm{T}_{current})$を求める.
  次に以下の工程を繰り返す.
  \begin{enumerate}
    \item $(\bm{\theta}_{calc}, \bm{T}_{calc}, \bm{l}_{calc}) = \bm{f}_{dec}(\bm{z})$を求める.
    \item 損失$L_{control}(\bm{\theta}_{calc}, \bm{T}_{calc}, \bm{l}_{calc})$を計算する.
    \item 誤差逆伝播\cite{rumelhart1986backprop}により$\bm{z}$を更新する.
  \end{enumerate}
  2)では, 以下のように損失$L_{control}$を計算する.
  \begin{align}
    L_{control}(\bm{\theta}, \bm{T}, \bm{l}) = &w_{1}||\bm{T}||_{2} + w_{2}||\bm{\theta}-\bm{\theta}_{target}||_{2}\nonumber\\ &+ w_{3}||\bm{\tau}_{target}+G^{T}(\bm{\theta}_{target}, \bm{T}_{current})\bm{T}||_{2} \label{eq:keep-control-loss}
  \end{align}
  ここで, $\bm{\tau}_{target}$は幾何モデルから計算された$\bm{\theta}_{target}$を保つのに必要な関節トルク値である.
  また, $G(\bm{\theta}, \bm{T})$は筋長ヤコビアンであり, 微小な変位$\Delta\bm{\theta}$を加えた$(\bm{\theta}+\Delta\bm{\theta}$, $\bm{T})$, または$(\bm{\theta}, \bm{T})$を\textcircled{\scriptsize1}の形でMusculoskeletal AutoEncoderに入力したときの出力$\bm{l}$の差から計算することができる.
  3)では, 以下のように誤差逆伝播法\cite{rumelhart1986backprop}を元に$\bm{z}$を最急降下法で更新する.
  \begin{align}
    \bm{g} &= dL_{control}/d\bm{z}\\
    \bm{z} &= \bm{z} - \gamma\bm{g}/|\bm{g}|
  \end{align}
  ここで, $\bm{g}$は$\bm{z}$に関する$L_{control}$の勾配である.
  このとき, $\gamma$の値を決め打ちしても良いが, 本研究では様々な$\gamma$によってバッチを作成し, 最も良い$\gamma$を選ぶ.
  $\gamma$の最大値$\gamma^{control}_{max}$を決め, 0から$\gamma^{control}_{max}$までの値を$N^{control}_{batch}$等分し, それらによって更新された$\bm{z}$を$N^{control}_{batch}$個作成する.
  もう一度1)と2)を行い, 最も$L_{control}$が小さかった$\bm{z}$を採用する.
  これら1)--3)の工程を, $C^{control}_{epoch}$回行う.
  そして最後に, 得られた$\bm{l}_{calc}$, $\bm{T}_{calc}$から以下のように$\bm{l}_{target}$を計算し, 実機に送る.
  \begin{align}
    \bm{l}_{target} = \bm{l}_{calc}+\bm{l}_{comp}(\bm{T}_{calc}) \label{eq:move-keep}
  \end{align}
  これにより, 関節角度$\bm{\theta}_{target}$を実現する中で, 筋張力が最小となるような$\bm{l}_{target}$を求めることが可能となる.
  $||\bm{T}||_{2}$を最小化し$\tau_{target}=-G^T\bm{T}$を満たすように二次計画法を解いて直接必要な筋張力$\bm{T}_{target}$を求めることも可能だが, $\bm{\theta}_{target}$において$\bm{T}_{target}$を実現できる保証はなく, 本研究では潜在空間$\bm{z}$内において探索している.

  最後に, 筋張力制御・可変剛性制御について説明する.
  筋張力制御は筋長ヤコビアンのみ必要なため, \cite{jantsch2011scalable, jantsch2012computed, kawamura2016jointspace}と同様に行うことができる.
  また, 可変剛性制御も同様に, \cite{kawaharazuka2019longtime}と同じように行うことが可能である.

  本研究では, $w_{1}=1.0$, $w_{2}=1.0$, $w_{3}=0.01$, $\gamma^{control}_{max}=0.5$, $C^{control}_{batch}=10$, $C^{control}_{epoch}=10$とする.
}%

\subsection{Simulation Using Musculoskeletal AutoEncoder} \label{subsec:simulation-method}
\switchlanguage%
{%
  We will explain the simulation method of the musculoskeletal humanoid using MAE.
  We simulate the transition of joint angles and muscle tensions by commanding muscle lengths.

  First, the state of latent space $\bm{z}=\bm{f}_{encode, 2}(\bm{T}_{sim}, \bm{l}_{sim})$ is calculated from the current muscle tension $\bm{T}_{sim}$ and length $\bm{l}_{sim}$ in the current simulation.
  Second, the procedures of 1)--3) are repeated as in \secref{subsec:control-method}.
  The difference is only the calculation of loss in 2).
  In the following explanations, we substitute $L_{sim, 1}$ or $L_{sim, 2}$ for the $L_{control}$ in \secref{subsec:control-method}.
  The loss of $L_{sim, 1}$ is calculated as shown below,
  \begin{align}
    L_{sim, 1}(\bm{\theta}, \bm{T}, \bm{l}) = &w_{4}||\bm{T}||_{2} + w_{5}||\bm{l}-\bm{l}_{sim}||_{2}\nonumber\\ &+ w_{6}||\bm{\tau}_{target}+G^{T}(\bm{\theta}_{sim}, \bm{T}_{sim})\bm{T}||_{2}\label{eq:torque-sim-loss}
  \end{align}
  By using this loss, the state of latent space $\bm{z}$, which fulfills the current muscle length and generates the joint torque $\bm{\tau}_{target}$ required for gravity compensation, can be obtained.
  When we would like to simulate motions when applying external force, we calculate the necessary torque from the geometric model, and conduct the simulation similarly.
  Also, by changing the loss $L$, we can simulate not only external force, but also motions when changing to a certain posture $\bm{\theta}_{fix}$ forcibly, when resting the arms on the table, etc.
  In this case, the loss of $L_{sim, 2}$ is calculated as shown below,
  \begin{align}
    L_{sim, 2}(\bm{\theta}, \bm{T}, \bm{l}) = w_{4}||\bm{T}||_{2} + w_{5}||\bm{l}-\bm{l}_{sim}||_{2} + w_{7}||\bm{\theta}-\bm{\theta}_{fix}||_{2}\label{eq:fix-sim-loss}
  \end{align}
  When using this loss, the simulator works more stably by changing the calculation of $\bm{z}$ to $\bm{z}=\bm{f}_{encode, 3}(\bm{\theta}_{sim}, \bm{l}_{sim})$.
  Finally, we obtain $(\bm{\theta}'_{sim}, \bm{T}'_{sim}, \bm{l}'_{sim}) = \bm{f}_{decode}(\bm{z})$ from the calculated $\bm{z}$, and update the simulator by setting $\bm{\theta}_{sim}=\bm{\theta}'_{sim}$ and $\bm{T}_{sim}=\bm{T}'_{sim}$.
  We can conduct simulation of the musculoskeletal humanoid by executing these procedures real-time.

  In this study, we set $w_{4}=0.1$, $w_{5}=1.0$, $w_{6}=0.001$, $w_{7}=1.0$, $\gamma^{sim}_{max}=0.2$, $C^{sim}_{batch}=10$, and $C^{sim}_{epoch}=3$
  $\gamma^{sim}_{max}$, $C^{sim}_{batch}$, and $C^{sim}_{epoch}$ mean $\gamma^{control}_{max}$, $C^{conrol}_{batch}$, and $C^{control}_{epoch}$ for simulation, respectively.
}%
{%
  Musculoskeletal AutoEncoderを用いた筋骨格ヒューマノイドのシミュレーションを行う.
  ここでは, 筋長を制御指令として関節角度と筋張力の遷移をシミュレーションしている.

  まず, 現在のシミュレーションにおける筋張力$\bm{T}_{sim}$, 筋長$\bm{l}_{sim}$から, 潜在状態$\bm{z}=\bm{f}_{encode, 2}(\bm{T}_{sim}, \bm{l}_{sim})$を求める.
  次に, \secref{subsec:control-method}と同様に, 1)--3)の工程を繰り返す.
  この工程のうち\secref{subsec:control-method}と異なるのは, 2)の損失の計算のみである.
  以降では, 前節の$L_{control}$を$L_{sim, 1}$または$L_{sim, 2}$で置き換える.
  2)では, 以下のように損失$L_{sim, 1}$を計算する.
  \begin{align}
    L_{sim, 1}(\bm{\theta}, \bm{T}, \bm{l}) = &w_{4}||\bm{T}||_{2} + w_{5}||\bm{l}-\bm{l}_{sim}||_{2}\nonumber\\ &+ w_{6}||\bm{\tau}_{target}+G^{T}(\bm{\theta}_{sim}, \bm{T}_{sim})\bm{T}||_{2}\label{eq:torque-sim-loss}
  \end{align}
  これにより, ロボットに送った筋長を満たしつつ, 重力補償等に必要なトルク$\bm{\tau}_{target}$を発揮する潜在空間$\bm{z}$を得ることができる.
  外力を加えた際の動きをシミュレーションしたい場合も, 関節の幾何モデルから必要なトルクを計算することができ, 同様にシミュレーションを行うことができる.
  また, 損失$L$を変えることによって, 外力だけでなく, 無理やり身体を外部からある姿勢$\bm{\theta}_{fix}$に動かした際やテーブル等に手を置いた時等の動きもシミュレーションすることが可能である.
  この場合の損失$L_{sim, 2}$は以下のようになる.
  \begin{align}
    L_{sim, 2}(\bm{\theta}, \bm{T}, \bm{l}) = w_{4}||\bm{T}||_{2} + w_{5}||\bm{l}-\bm{l}_{sim}||_{2} + w_{7}||\bm{\theta}-\bm{\theta}_{fix}||_{2}\label{eq:fix-sim-loss}
  \end{align}
  注意点として, 損失関数をこの$L_{sim, 2}$にする場合は, 最初の潜在空間$\bm{z}$を求める際に$\bm{z}=\bm{f}_{encode, 3}(\bm{\theta}_{sim}, \bm{l}_{sim})$とした方がシミュレータが安定して動作した.
  最後に, ここで最終的に求まった$\bm{z}$から, $(\bm{\theta}'_{sim}, \bm{T}'_{sim}, \bm{l}'_{sim}) = \bm{f}_{decode}(\bm{z})$を求め, $\bm{\theta}_{sim}=\bm{\theta}'_{sim}$, $\bm{T}_{sim}=\bm{T}'_{sim}$のようにシミュレータを更新する.
  これら全ての工程をリアルタイムで回すことで, シミュレーションが行われる.

  本研究では, $w_{4}=0.1$, $w_{5}=1.0$, $w_{6}=0.001$, $w_{7}=1.0$, $\gamma^{sim}_{max}=0.2$, $C^{sim}_{batch}=10$, $C^{sim}_{epoch}=3$とする.
  ここで, $\gamma^{sim}_{max}, C^{sim}_{batch}, C^{sim}_{epoch}$はそれぞれ, シミュレーション内の計算における$\gamma^{control}_{max}, C^{conrol}_{batch}, C^{control}_{epoch}$と同等の定数を指す.
}%

\section{Experiments} \label{sec:experiments}
\switchlanguage%
{%
  As stated in \secref{subsec:musculoskeletal-humanoids}, we mainly use the 5 DOFs joints of the shoulder and elbow for the evaluation of this study, and so we set $D=5$ and $M=10$.
}%
{%
  本研究では, \secref{subsec:musculoskeletal-humanoids}で説明したように, 基本的に肩の5自由度に関して実験し評価を行うため, $D=5$, $M=10$とする.
}%

% \begin{figure}[t]
%   \centering
%   \includegraphics[width=1.0\columnwidth]{figs/initial-training}
%   \caption{The result of initial training of Musculoskeletal AutoEncoder.}
%   \label{figure:initial-training}
%   % \vspace{-1.0zh}
% \end{figure}

\begin{figure}[t]
  \centering
  \includegraphics[width=0.9\columnwidth]{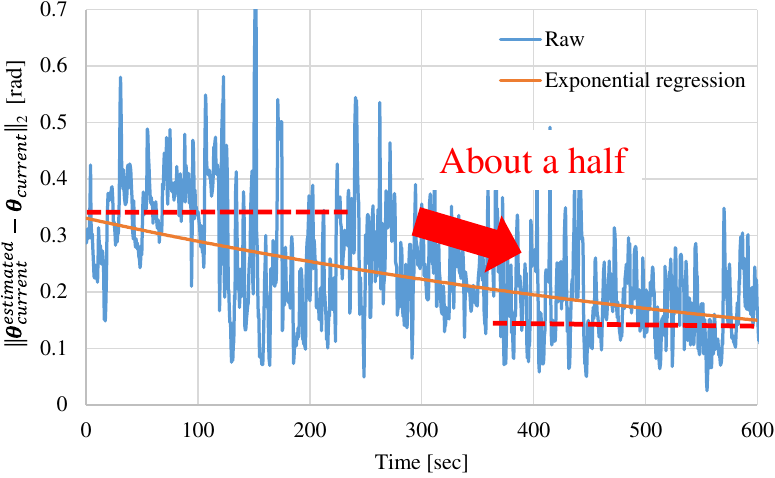}
  \caption{The transition of the error between the current and estimated joint angles while executing online learning of Musculoskeletal AutoEncoder.}
  \label{figure:online-learning}
  \vspace{-1.0ex}
\end{figure}

\begin{figure}[t]
  \centering
  \includegraphics[width=1.0\columnwidth]{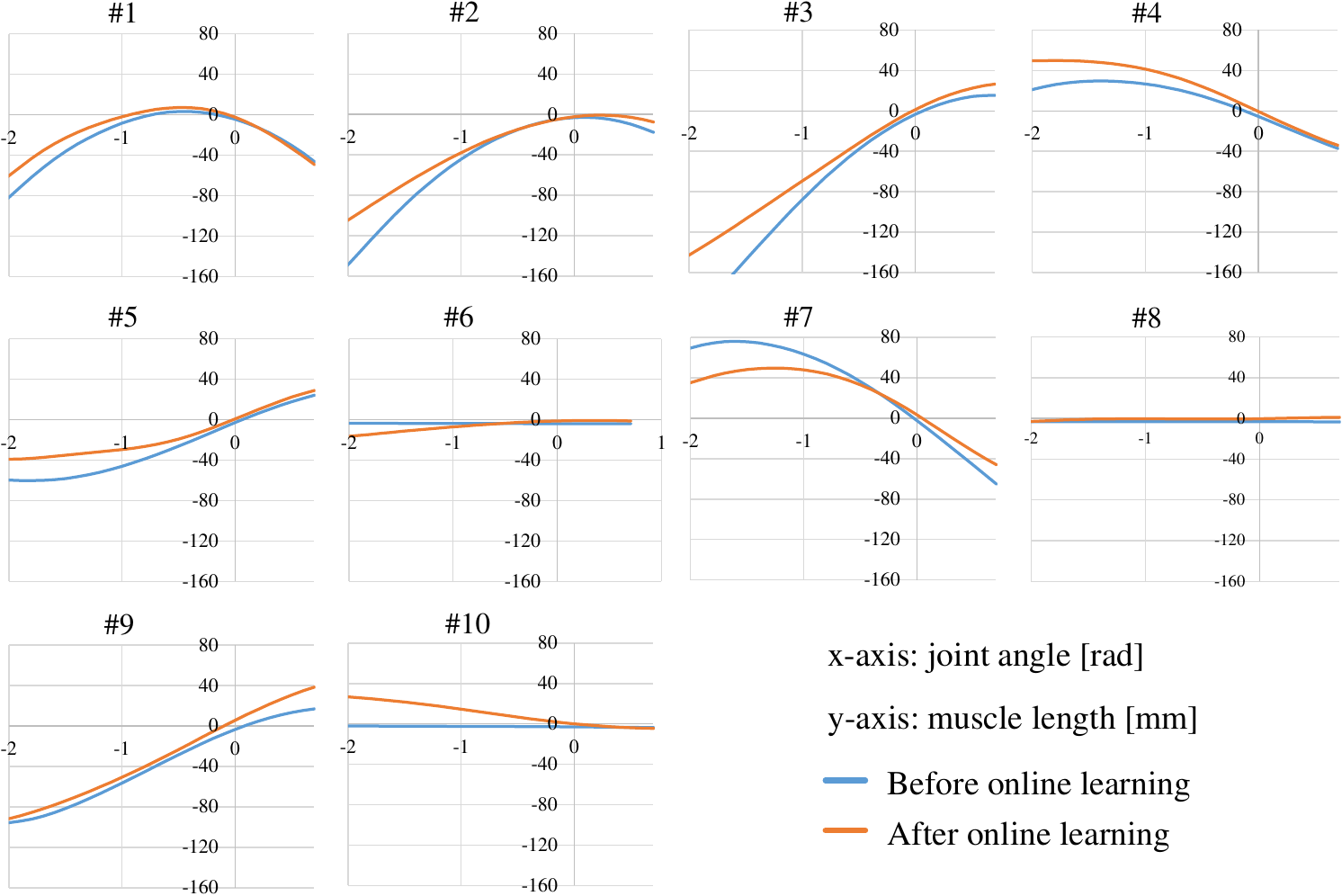}
  \caption{The transition of muscle lengths by changing joint angles in Musculoskeletal AutoEncoder, before and after the online learning.}
  \label{figure:show-model}
  \vspace{-3.0ex}
\end{figure}

\begin{figure}[t]
  \centering
  \includegraphics[width=1.0\columnwidth]{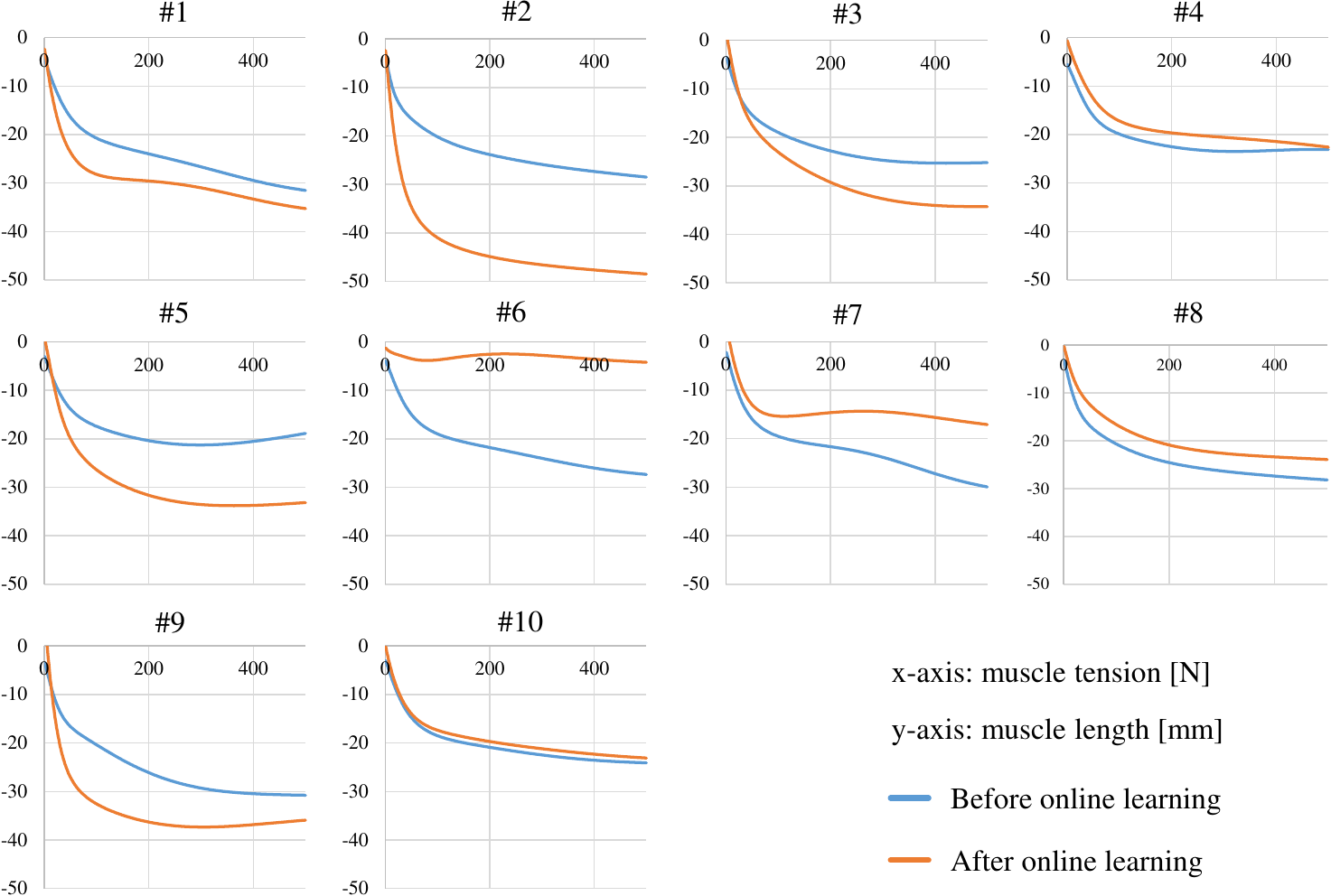}
  \caption{The transition of muscle lengths by changing muscle tensions in Musculoskeletal AutoEncoder, before and after the online learning.}
  \label{figure:show-emodel}
  \vspace{-3.0ex}
\end{figure}

\subsection{Online Learning of Musculoskeletal AutoEncoder} \label{subsec:training-experiment}
\switchlanguage%
{%
  We conducted an experiment regarding the online learning of MAE.
  In this study, we moved the robot by 3 steps as in \cite{kawaharazuka2019longtime} and updated MAE.
  First, we conducted \equref{eq:move-first} over 5 sec by setting target joint angle $\bm{\theta}_{target}$ randomly and setting $\bm{T}_{const}$ as $\bm{T}_{bias}$.
  Second, we conducted \equref{eq:move-second} over 2 sec with the same $\bm{\theta}_{target}$.
  Third, we conducted \equref{eq:move-first} over 3 sec again by setting $\bm{T}_{const}$ randomly in the range of $[\bm{T}_{bias}, \bm{T}_{limit}]$.
  By repeating these 3 steps, the multidimensional space of joint angles and muscle tensions in MAE is efficiently updated.
  In these steps, we set $\bm{T}_{bias}= 30$ [N] and $\bm{T}_{limit}= 200$ [N].

  We show the transition of the difference between the estimated joint angles $\bm{\theta}^{estimated}_{current}$ and values of joint angle sensors $\bm{\theta}_{current}$, $||\bm{\theta}^{estimated}_{current}-\bm{\theta}_{current}||_{2}$, in \figref{figure:online-learning}.
  During online learning, $||\bm{\theta}^{estimated}_{current}-\bm{\theta}_{current}||_{2}$ gradually decreased.
  From the exponential approximation of 10 minutes, we can see that the error of joint angle estimation decreased by about half after 10 minutes, from 0.324 [rad] to 0.154 [rad] on average.
  The error of $||\bm{\theta}^{estimated}_{current}-\bm{\theta}_{current}||_{2}$ remained, and it is considered to be due to muscle hysteresis caused by friction of joints and muscles.
  This study can only consider quasi-static movements and cannot consider dynamics-related errors such as hysteresis.

  We show the change of MAE before and after the online learning in \figref{figure:show-model} and \figref{figure:show-emodel}.
  In \figref{figure:show-model}, we show the change in muscle lengths when setting muscle tension as $\bm{0}$ and changing $\bm{\theta}_{S-p}$ in the range of $[-120, 30]$ [deg].
  In \figref{figure:show-emodel}, we show the change in muscle lengths when changing muscle tension in the range of $[0, 500]$ [N] at the initial posture.
  Both of the graphs change largely after online learning.
  Muscle lengths in \figref{figure:show-model} changed up to about 40 [mm] while passing the origin.
  Especially, the changes in muscle lengths of the major muscles that move $\bm{\theta}_{S-p}$, e.g. $\#2$, $\#3$, and $\#7$ in \figref{figure:musculoskeletal-humanoid}, are large.
  Also, muscles in \figref{figure:show-emodel} expressed various characteristics of linearity, e.g. $\#2$, and nonlinearity, e.g. $\#6$.
  Because $\#2$ is a muscle which large tension is often applied to, the characteristic of the nonlinear elastic unit changed largely.
  Also, because $\#6$ is a muscle of the elbow joint and the friction is large due to its muscle route, the nonlinearity vanished.
}%
{%
  % まず, Musculoskeletal AutoEncoderの初期学習について述べる.
  % 結果を\figref{figure:initial-training}に示す.
  % $L_{\bm{\theta}}=||\bm{\theta}-\bm{\theta}_{calc}||^2$, $L_{\bm{T}}=||\bm{T}-\bm{T}_{calc}||^2$, $L_{\bm{l}}=||\bm{l}-\bm{l}_{calc}||^2$とし, $\cdot^{val}$はtestデータを用いたときの損失を表す.
  % train, testともに正しく損失が下がっていることがわかる.
  Musculoskeletal AutoEncoderのオンライン学習について述べる.
  本研究では\cite{kawaharazuka2019longtime}と同様に, 3段階でロボットを動かすことを繰り返し, 学習を行う.
  まず, \equref{eq:move-first}において関節角度限界の中でランダムに$\bm{\theta}_{target}$を指定し, $\bm{T}_{const}$は最初に一律で$\bm{T}_{bias}$を指定する.
  次に指令関節角度をそのままで\equref{eq:move-second}を実行し, 最後に, 筋張力指令を$\bm{T}_{bias}$から$\bm{T}_{limit}$の間でランダムに指定してもう一度\equref{eq:move-first}を実行する.
  この3つの手順を繰り返すことで, 効率よく関節角度と筋張力の空間を学習させる.
  ここで, $\bm{T}_{bias}= 30$ [N], $\bm{T}_{limit}= 200$ [N]とする.
  関節角度推定値$\bm{\theta}^{estimated}_{current}$と関節モジュールのセンサ値$\bm{\theta}_{current}$の差分$||\bm{\theta}^{estimated}_{current}-\bm{\theta}_{current}||_{2}$の遷移を\figref{figure:online-learning}に示す.
  オンライン学習をするにつれ, $||\bm{\theta}^{estimated}_{current}-\bm{\theta}_{current}||_{2}$が徐々に下がっていくことがわかる.
  10分間の実験を行った結果を指数関数近似した曲線から, 関節角度推定誤差は10分間で0.324 [rad]から0.154 [rad]と, 約半分程度まで下がっていることがわかる.

  オンライン学習前と学習後におけるMusculoskeletal AutoEncoderの変化を\figref{figure:show-model}, \figref{figure:show-emodel}に示す.
  \figref{figure:show-model}は筋張力が0の状態で, $\bm{\theta}_{S-p}$を$[-120, 30]$ [deg]の範囲で変化させたときの筋長変化を表している.
  また, \figref{figure:show-emodel}は初期姿勢において筋張力を$[0, 500]$ [N]の範囲で変化させたときの筋長変化を表している.
  両者とも, 学習前と学習後でそれぞれ大きく変化している.
  \figref{figure:show-model}では, 関節変化に対する筋変化は原点を維持しながら最大で40 [mm]程度変化している.
  特に, $\bm{\theta}_{S-p}$を動作させる主要な筋$\#2, \#3, \#7$に大きな変化が起きていることもわかる.
  また, \figref{figure:show-emodel}では, より大きく非線形性が出るもの($\#2$)やほぼ直線になるも$\#6$のまで様々である.
  $\#2$は非常に強い力がかかりやすい筋であり, 非線形弾性要素が劣化で変化したため非線形性がより大きく出ていると考えられる.
  また, $\#6$は肘に関わる筋で取り回しの都合上非常に摩擦が強く, 非線形性が消えてしまっていると考えられる.

  最後に, $||\bm{\theta}^{estimated}_{current}-\bm{\theta}_{current}||_{2}$に誤差が残ってしまうのは, 筋や関節の摩擦によるヒステリシスの影響が大きいと考えられる.
}%

\subsection{State Estimation Using Musculoskeletal AutoEncoder} \label{subsec:estimation-experiment}
\begin{figure}[t]
  \centering
  \includegraphics[width=1.0\columnwidth]{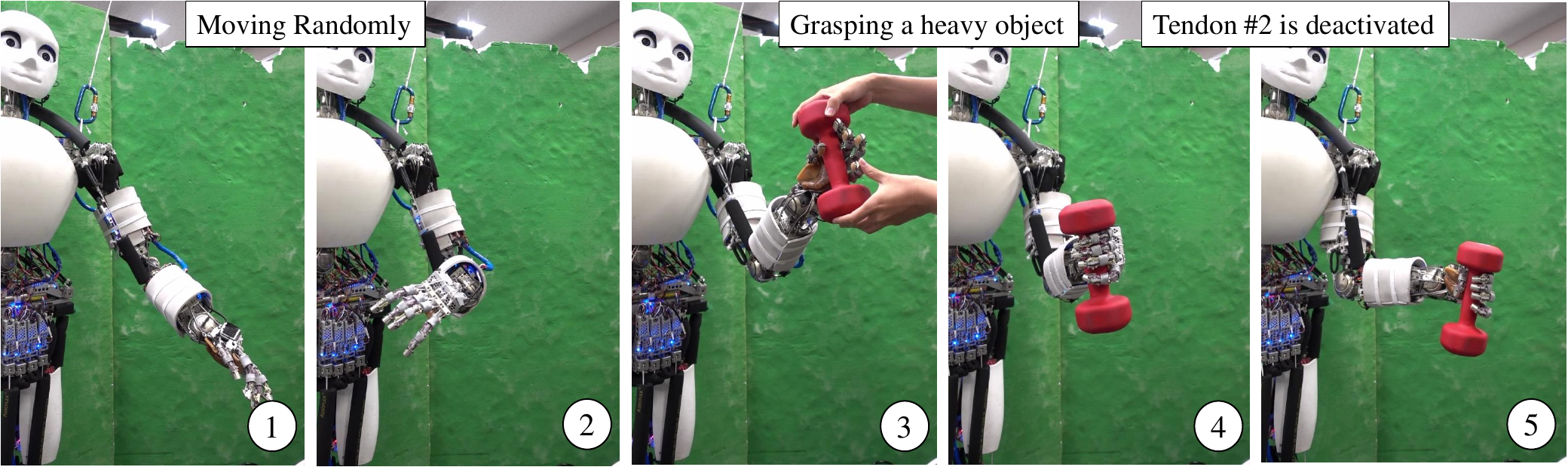}
  \caption{The experiment of state estimation using Musculoskeletal AutoEncoder.}
  \label{figure:state-estimation-appearance}
  \vspace{-1.0ex}
\end{figure}

\begin{figure}[t]
  \centering
  \includegraphics[width=1.0\columnwidth]{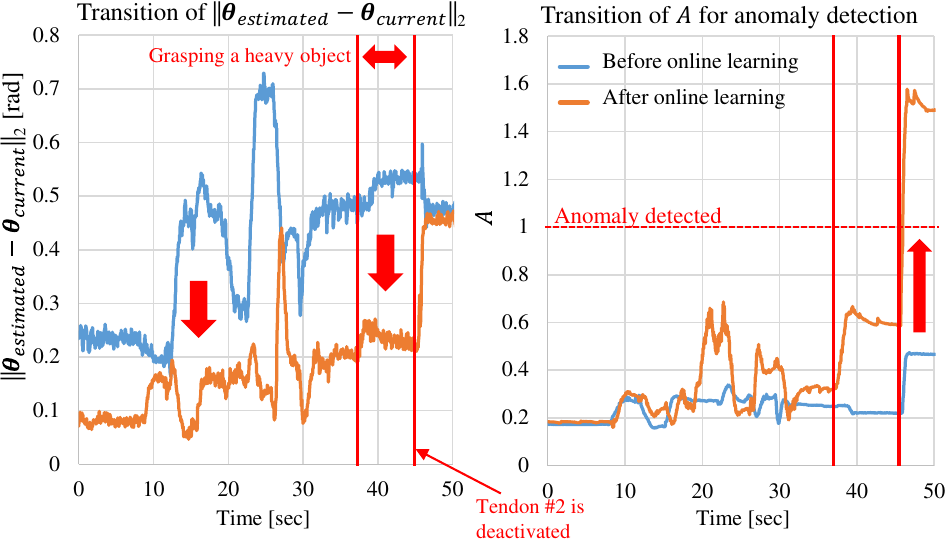}
  \caption{The result of state estimation using Musculoskeletal AutoEncoder.}
  \label{figure:state-estimation}
  \vspace{-3.0ex}
\end{figure}

\switchlanguage%
{%
  We conducted an experiment of state estimation using MAE.
  We sent the same motions to the robot before and after online learning, .
  In the middle of the experiment, the robot grasped a heavy object (3.6 kg dumbbell), and then we deactivated the muscle $\#2$ shown in \figref{figure:musculoskeletal-humanoid}.
  We evaluated the tracking ability of the estimated joint angles and the value $A$ for anomaly detection before and after online learning.
  We did not conduct online learning during the motions.
  We show the experiment in \figref{figure:state-estimation-appearance}, and its result in \figref{figure:state-estimation}.

  When moving randomly, the error of the estimated joint angles largely decreased after online learning from 0.414 [rad] to 0.186 [rad] on average.
  The space of muscle tension is efficiently learned during online learning, and the error did not change significantly when grasping the heavy dumbbell.
  In the case before online learning, the change in error also seems small because the error was originally high.
  Next, we will examine the transition of $A$.
  After online learning, $A$ rapidly rose when deactivating muscle $\#2$ and we can detect anomaly well.
  On the other hand, $A$ did not rise very much before online learning.
  This is considered to be because we set muscle tension randomly at initial training, and MAE is trained to restore the input value even when setting muscle tension impossible at that posture.
}%
{%
  状態推定について実機実験を行い評価する.
  オンライン学習を行った後にそれを止め, 学習する前とした後における関節角度推定値の追従の差を評価した.
  実験の途中で, 重量物体の把持, また, 筋一本の機能を停止することを行い, その際の関節角度推定値の追従, また, 異常度$A$の測定を行う.
  本実験中は一切のオンライン学習は行っていない.
  実験の様子を\figref{figure:state-estimation-appearance}に, 実験結果を\figref{figure:state-estimation}に示す.

  ランダムに関節を動かした際には, 学習前と後で, 関節角度推定誤差が, 平均で0.414 [rad]から0.186 [rad]と, 大きく下がっていることがわかる.
  また, オンライン学習の際に筋張力の空間も効率的に学習しているため, 重量物体を把持した際も大きく関節角度推定誤差が上がることはなかった.
  % TODO (計算スピードについて評価もしてみたい, table付きで, muscle jacobianを抜かして)
  最後に, 異常検知度$A$の推移を観察する.
  学習後は, 一本の筋(\figref{figure:musculoskeletal-humanoid}の$\#2$)の機能を停止することで, $A$が急上昇しており, 異常を検知可能なことがわかった.
  一方で, 学習前では$A$はあまり上昇しない.
  これは, 初期学習の際には筋張力を全空間でランダムに決めているため, 実行不可能な筋張力においても同じようにセンサ値を復元可能であり, $A$が下がるためである.
}%

\subsection{Control Using Musculoskeletal AutoEncoder} \label{subsec:control-experiment}
\begin{figure}[t]
  \centering
  \includegraphics[width=1.0\columnwidth]{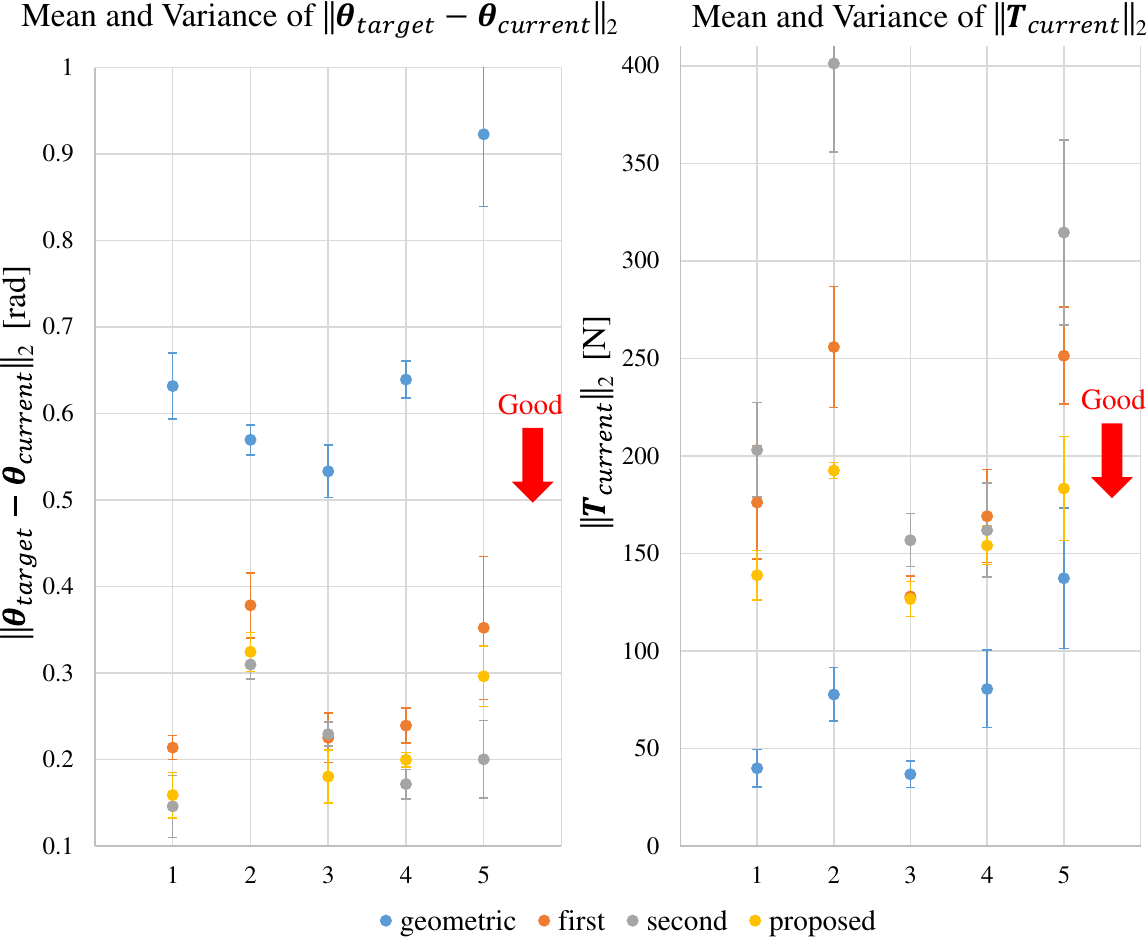}
  \caption{The average and variance of $||\bm{\theta}_{eval}-\bm{\theta}_{current}||_{2}$ and $||\bm{T}_{current}||_{2}$ among four controls: \textbf{geometric}, \textbf{first}, \textbf{second}, and \textbf{proposed}.}
  \label{figure:control-experiment}
  \vspace{-3.0ex}
\end{figure}

\switchlanguage%
{%
  We conducted experiments regarding joint angle control using MAE.
  First, we decided 5 random joint angles $\bm{\theta}_{eval}$ for evaluation.
  We moved the robot from the random joint angle $\bm{\theta}_{random}$ to $\bm{\theta}_{eval}$ 5 times while changing $\bm{\theta}_{random}$, and evaluated the average and variance of $||\bm{\theta}_{eval}-\bm{\theta}_{current}||_{2}$ and $||\bm{T}_{current}||_{2}$.
  We conducted the experiment in this way to consider the effect of muscle hysteresis.
  We show the result of moving the robot using the geometric model (\textbf{geometric}), using \equref{eq:move-first} (\textbf{first}) and \equref{eq:move-second} (\textbf{second}) in order after online learning, and using \equref{eq:move-keep} after online learning (\textbf{proposed}) in \figref{figure:control-experiment}.

  From the result, we can see that the precision using controls after online learning (\textbf{first}, \textbf{second}, and \textbf{proposed}) were much better than that using the geometric model (\textbf{geometric}).
  Also, the joint angle error of \textbf{second} was better than \textbf{first}, because \textbf{second} uses the current muscle tension.
  The joint angle error of \textbf{proposed} was almost the same with \textbf{second}.
  Muscle tension of \textbf{geometric} was much lower than those using other methods.
  This is considered to be due to the loosening of muscles, since the difference between the actual robot and its geometric model is large and the control using the geometric model could not realize target joint angles.
  Muscle tension was likely to rise in \textbf{second} compared to in \textbf{first}.
  On the other hand, $||\bm{T}_{current}||_{2}$ in \textbf{proposed} was lower than \textbf{first} and \textbf{second} at all $\bm{\theta}_{eval}$.

  Because \textbf{first} and \textbf{second} are executed in order, it takes a long time to realize the target joint angles, and high internal muscle tension emerges.
  On the other hand, \textbf{proposed} can inhibit internal muscle tension and can realize the same tracking ability as \textbf{second}.
}%
{%
  筋長による関節位置制御について実機実験を行い評価する.
  まず, 評価を行う関節角度$\bm{\theta}_{eval}$を5点ランダムに決定する.
  あるランダムな関節角度$\bm{\theta}_{random}$から$\bm{\theta}_{eval}$に動作することを$\bm{\theta}_{random}$を変えながら5回行い, そのときの$||\bm{\theta}_{eval}-\bm{\theta}_{current}||_{2}, ||\bm{T}_{current}||_{2}$の平均と分散を評価する.
  これは, ヒステリシス等の影響を考慮してこのように行った.
  幾何モデルを直接使った際(geometric), 学習後に\equref{eq:move-first}, \equref{eq:move-second}の順に動かした際の結果(first, second), また, 学習後に\equref{eq:move-keep}を用いて動かした際(keep)の結果を\figref{figure:control-experiment}に示す.

  幾何モデルよりも, 学習後の制御の方が圧倒的に関節角度の実現精度が高いことがわかる.
  また, \equref{eq:move-second}は\equref{eq:move-first}の後の現在筋張力を使うため, より精度が高い.
  \equref{eq:move-keep}は精度に関しては\equref{eq:move-second}とほぼ同じような挙動をしている.
  また, 筋張力に関しては幾何モデルを使ったものが他の手法に比べ圧倒的に小さい.
  これは, 学習されていないため幾何モデルと実機の誤差が大きいため, 指令関節角度を実現できず緩んでしまっているからであると考えられる.
  また, \equref{eq:move-second}は筋張力が高まりやすく, \equref{eq:move-first}に比べて高い筋張力を発揮してしまっていることがわかる.
  それに対して, \equref{eq:move-keep}は全姿勢において\equref{eq:move-first}や\equref{eq:move-second}よりも$||\bm{T}_{current}||_{2}$が小さい.

  \equref{eq:move-first}, \equref{eq:move-second}は順に実行しなければならないため姿勢の実現に時間がかかるかつ内力の高まりが大きい.
  それに対して, \equref{eq:move-keep}は内力の高まりを抑えつつ, \equref{eq:move-second}と同等の追従性を得ることが出来た.
}%

\begin{figure}[t]
  \centering
  \includegraphics[width=0.9\columnwidth]{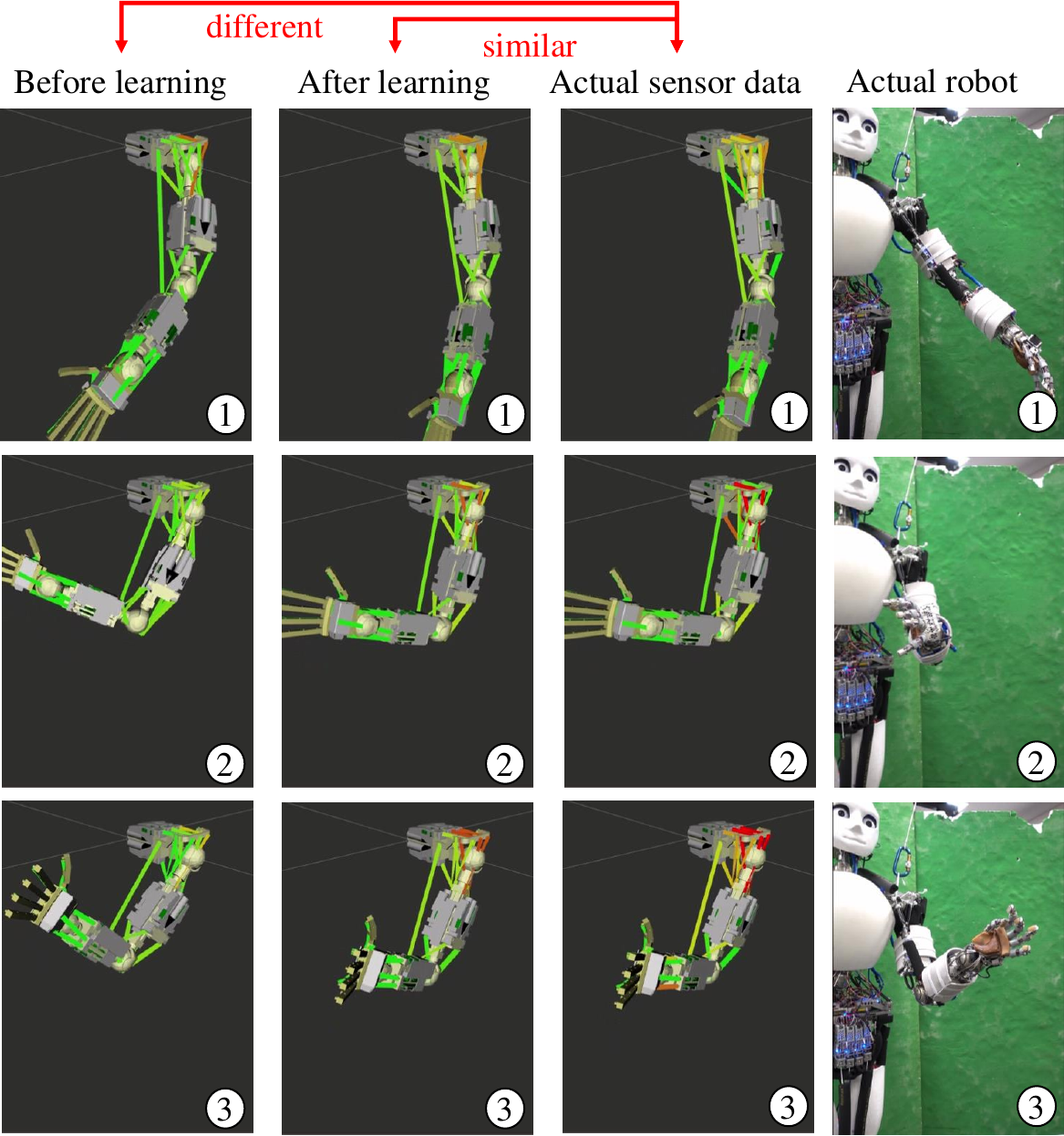}
  \caption{The differences among the actual robot, simulation before online learning, and simulation after online learning.}
  \label{figure:simulation-experiment}
  \vspace{-1.0ex}
\end{figure}

\begin{figure}[t]
  \centering
  \includegraphics[width=1.0\columnwidth]{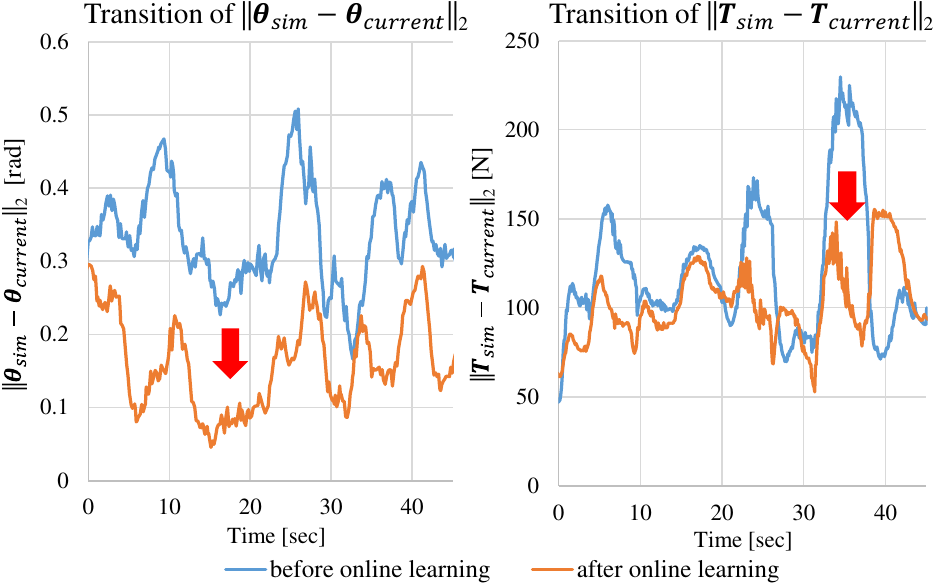}
  \vspace{-3.0ex}
  \caption{The transition of $||\bm{\theta}_{sim}-\bm{\theta}_{current}||_{2}$ and $||\bm{T}_{sim}-\bm{T}_{current}||_{2}$ before and after online learning.}
  \label{figure:simulation-graph}
  \vspace{-3.0ex}
\end{figure}

\subsection{Simulation Using Musculoskeletal AutoEncoder} \label{subsec:simulation-experiment}
\begin{figure}[t]
  \centering
  \includegraphics[width=1.0\columnwidth]{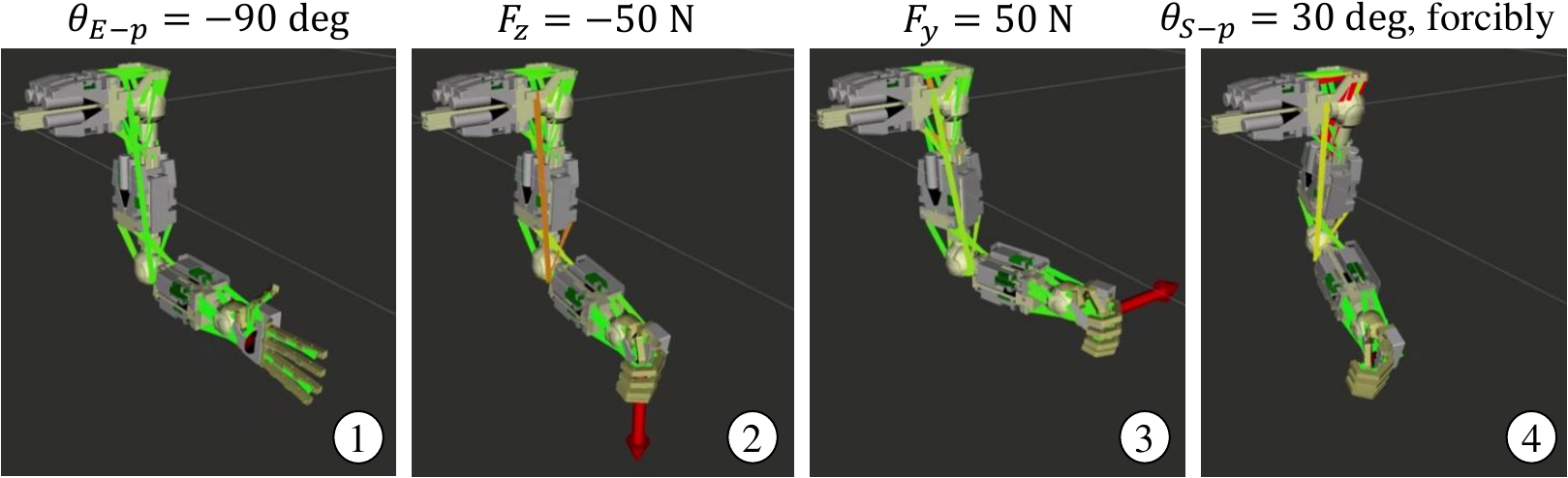}
  \vspace{-3.0ex}
  \caption{The result of simulation experiment when changing the loss function.}
  \label{figure:simulation-change}
  \vspace{-3.0ex}
\end{figure}

\switchlanguage%
{%
  We conducted experiments of simulation using MAE.
  We show the difference of movements among the actual robot, simulation before online learning, and simulation after online learning in \figref{figure:simulation-experiment}.
  The higher the muscle tension is, the redder its color becomes, and the lower the muscle tension is, the greener its color becomes.
  Compared to before online learning, we can see that the actual sensor values of joint angles and muscle tensions are close to the simulation after online learning.
  We show the transition of $||\bm{\theta}_{sim}-\bm{\theta}_{current}||_{2}$ and $||\bm{T}_{sim}-\bm{T}_{current}||_{2}$ before and after online learning in \figref{figure:simulation-graph}.
  We can see that the error between the actual sensor value and simulation value becomes smaller after online learning than before online learning, e.g. regarding joint angles, from 0.335 [rad] to 0.162 [rad] on average, and regarding muscle tensions, from 105 [N] to 92.2 [N] on average.
  Thus, we can make the simulator closer to the actual robot by learning.

  Also, we show that the behavior of simulation can be altered by changing loss function, as stated in \secref{subsec:simulation-method}, in \figref{figure:simulation-change}.
  We conducted the movements of bending the elbow to 90 deg, applying external force of -50 N in $z$ direction and 50 N in $y$ direction in order, and changing $\theta_{S-p}$ to 30 deg forcibly by external force.
  When applying external force, we use the loss of \equref{eq:torque-sim-loss}, and when changing the posture forcibly, we use the loss of \equref{eq:fix-sim-loss}.
  By changing the applied force or setting $\bm{\theta}_{fix}$ in \equref{eq:fix-sim-loss}, we can check the transitions of joint angles and muscle tensions.
  In the future, we would like to make use of the simulator for motion generation.
}%
{%
  シミュレーションに関する実験を行う.
  オンライン学習前と学習後において, 実機の動きとシミュレータ上のロボットの動きの差異を評価する.
  学習前のシミュレータ・学習後のシミュレータ・実機動作の際におけるセンサ値の比較を\figref{figure:simulation-experiment}に示す.
  筋張力の値が高いほど筋の色が赤く, 小さいほど色が緑色になるようにしている.
  学習前に比べ, 学習後の方が実機センサ値に筋張力・関節角度が近づいていることがわかる.
  学習前と学習後における$||\bm{\theta}_{sim}-\bm{\theta}_{current}||_{2}$と$||\bm{T}_{sim}-\bm{T}_{current}||_{2}$の遷移を\figref{figure:simulation-graph}に示す.
  学習前に比べ学習後の方が, 関節角度については平均で0.335 [rad]から0.162 [rad], 筋張力については平均で104.8 [N]から92.2 [N]と, 実機センサ値とシミュレータ値の誤差が小さくなっていることがわかる.
  これにより, シミュレータの挙動を学習により実機に近づけていくことが可能であることがわかった.

  また, \secref{subsec:simulation-method}に示したように, 損失関数を変えることでシミュレータの挙動を変化させることができる様子を\figref{figure:simulation-change}に示す.
  これは, 肘を90度に曲げ, $z$方向に-50 N, $y$方向に50 Nを順にかけ, その後$\theta_{S-p}$を無理やり30 degにした際の挙動である.
  力を指定する際は\equref{eq:torque-sim-loss}を用い, 姿勢を指定する際は\equref{eq:fix-sim-loss}を使用している.
  かける力を変えたり, $\bm{\theta}_{fix}$を与えることで, それに応じた関節角度・筋張力を確認することが可能であり, 今後これらシミュレータを動作生成に有効活用していきたい.
}%

\section{CONCLUSION} \label{sec:conclusion}
\switchlanguage%
{%
  In this study, we proposed Musculoskeletal AutoEncoder (MAE) that uniformly handles state estimation, control, and simulation of musculoskeletal humanoids.
  MAE is an AutoEncoder-type network representing the relationship among joint angles, muscle lengths, and muscle tensions, which are equipped with the ordinary musculoskeletal humanoid, and we update it online using the actual robot sensor information.
  For state estimation, we can estimate the current joint angle by feeding the current muscle tension and length into MAE, and can also detect anomaly from whether the input values can be restored by MAE.
  For muscle length-based control, we can obtain the target muscle length to realize the target joint angle and minimize muscle tension by backpropagation of MAE.
  For simulation of joint angles and muscle tensions, we can calculate them to fulfill the necessary muscle length and joint torque by real-time backpropagation.
  Also, we can conduct muscle tension-based control and variable stiffness control by calculating muscle Jacobian from MAE.
  By using this study, we can integrate indispensable system components of musculoskeletal humanoids, and reflect the actual robot information to the model.

  In future works, we would like to investigate the control and simulation techniques of musculoskeletal humanoids for dynamic movements.
}%
{%
  本研究では, 筋骨格ヒューマノイドにおける状態推定・制御・シミュレーションを統一的に扱うMusculoskeletal AutoEncoderについて提案した.
  筋骨格ヒューマノイドに一般的に備わる関節角度・筋長・筋張力センサの関係をAutoEncoder型のネットワークで表現し, これを実機センサ情報からオンラインで更新していく.
  関節角度の状態推定は, 現在の筋長・筋張力を本ネットワークに通すことで行うことができる.
  関節の位置制御は, 指令関節角度と筋張力最小化の観点から誤差逆伝播を通して指令筋長を求め, 実行することができる.
  関節角度と筋張力のシミュレーションは, 指令筋長と必要関節トルクの観点から誤差逆伝播を通してリアルタイムに行うことができる.
  また, Musculoskeletal AutoEncoderの微分により筋長ヤコビアンを求めることで, 関節トルク制御, 可変剛性制御も同様に実行することができる.
  本研究により, 筋骨格ヒューマノイドの静的な動作に関する必須の要素群を統合し, かつ実機情報をオンラインでモデルに取り込んでいくことが可能となった.

  今後は, 動的な動作に関する制御やシミュレーション技術の開発を行っていきたい.
}%

{
  %\footnotesize
  %\small
  %\bibliographystyle{junsrt}
  \bibliographystyle{IEEEtran}
  \bibliography{main}

\begin{thebibliography}{10}
\providecommand{\url}[1]{#1}
\csname url@rmstyle\endcsname
\providecommand{\newblock}{\relax}
\providecommand{\bibinfo}[2]{#2}
\providecommand\BIBentrySTDinterwordspacing{\spaceskip=0pt\relax}
\providecommand\BIBentryALTinterwordstretchfactor{4}
\providecommand\BIBentryALTinterwordspacing{\spaceskip=\fontdimen2\font plus
\BIBentryALTinterwordstretchfactor\fontdimen3\font minus
  \fontdimen4\font\relax}
\providecommand\BIBforeignlanguage[2]{{%
\expandafter\ifx\csname l@#1\endcsname\relax
\typeout{** WARNING: IEEEtran.bst: No hyphenation pattern has been}%
\typeout{** loaded for the language `#1'. Using the pattern for}%
\typeout{** the default language instead.}%
\else
\language=\csname l@#1\endcsname
\fi
#2}}

\bibitem{nakanishi2013design}
Y.~Nakanishi, S.~Ohta, T.~Shirai, Y.~Asano, T.~Kozuki, Y.~Kakehashi,
  H.~Mizoguchi, T.~Kurotobi, Y.~Motegi, K.~Sasabuchi, J.~Urata, K.~Okada,
  I.~Mizuuchi, and M.~Inaba, ``{Design Approach of Biologically-Inspired
  Musculoskeletal Humanoids},'' \emph{International Journal of Advanced Robotic
  Systems}, vol.~10, no.~4, pp. 216--228, 2013.

\bibitem{wittmeier2013toward}
S.~Wittmeier, C.~Alessandro, N.~Bascarevic, K.~Dalamagkidis, D.~Devereux,
  A.~Diamond, M.~J{\"a}ntsch, K.~Jovanovic, R.~Knight, H.~G. Marques,
  P.~Milosavljevic, B.~Mitra, B.~Svetozarevic, V.~Potkonjak, R.~Pfeifer,
  A.~Knoll, and O.~Holland, ``{Toward Anthropomimetic Robotics: Development,
  Simulation, and Control of a Musculoskeletal Torso},'' \emph{Artificial
  Life}, vol.~19, no.~1, pp. 171--193, 2013.

\bibitem{jantsch2013anthrob}
M.~J{\"a}ntsch, S.~Wittmeier, K.~Dalamagkidis, A.~Panos, F.~Volkart, and
  A.~Knoll, ``{Anthrob - A Printed Anthropomimetic Robot},'' in
  \emph{Proceedings of the 2013 IEEE-RAS International Conference on Humanoid
  Robots}, 2013, pp. 342--347.

\bibitem{asano2016kengoro}
Y.~Asano, T.~Kozuki, S.~Ookubo, M.~Kawamura, S.~Nakashima, T.~Katayama,
  Y.~Iori, H.~Toshinori, K.~Kawaharazuka, S.~Makino, Y.~Kakiuchi, K.~Okada, and
  M.~Inaba, ``{Human Mimetic Musculoskeletal Humanoid Kengoro toward Real World
  Physically Interactive Actions},'' in \emph{Proceedings of the 2016 IEEE-RAS
  International Conference on Humanoid Robots}, 2016, pp. 876--883.

\bibitem{ookubo2015learning}
S.~Ookubo, Y.~Asano, T.~Kozuki, T.~Shirai, K.~Okada, and M.~Inaba, ``{Learning
  Nonlinear Muscle-Joint State Mapping Toward Geometric Model-Free Tendon
  Driven Musculoskeletal Robots},'' in \emph{Proceedings of the 2015 IEEE-RAS
  International Conference on Humanoid Robots}, 2015, pp. 765--770.

\bibitem{nakanishi2010estimation}
Y.~Nakanishi, K.~Hongo, I.~Mizuuchi, and M.~Inaba, ``{Joint proprioception
  acquisition strategy based on joints-muscles topological maps for
  musculoskeletal humanoids},'' in \emph{Proceedings of the 2010 IEEE
  International Conference on Robotics and Automation}, 2010, pp. 1727--1732.

\bibitem{motegi2012jacobian}
Y.~Motegi, T.~Shirai, T.~Izawa, T.~Kurotobi, J.~Urata, Y.~Nakanishi, K.~Okada,
  and M.~Inaba, ``{Motion control based on modification of the Jacobian map
  between the muscle space and work space with musculoskeletal humanoid},'' in
  \emph{Proceedings of the 2012 IEEE-RAS International Conference on Humanoid
  Robots}, 2012, pp. 835--840.

\bibitem{mizuuchi2006acquisition}
I.~Mizuuchi, Y.~Nakanishi, T.~Yoshikai, M.~Inaba, H.~Inoue, and O.~Khatib,
  ``{Body Information Acquisition System of Redundant Musculo-Skeletal
  Humanoid},'' in \emph{Experimental Robotics IX}, 2006, pp. 249--258.

\bibitem{kawaharazuka2018online}
K.~Kawaharazuka, S.~Makino, M.~Kawamura, Y.~Asano, K.~Okada, and M.~Inaba,
  ``{Online Learning of Joint-Muscle Mapping using Vision in Tendon-driven
  Musculoskeletal Humanoids},'' \emph{IEEE Robotics and Automation Letters},
  vol.~3, no.~2, pp. 772--779, 2018.

\bibitem{kawaharazuka2018bodyimage}
K.~Kawaharazuka, S.~Makino, M.~Kawamura, A.~Fujii, Y.~Asano, K.~Okada, and
  M.~Inaba, ``{Online Self-body Image Acquisition Considering Changes in Muscle
  Routes Caused by Softness of Body Tissue for Tendon-driven Musculoskeletal
  Humanoids},'' in \emph{Proceedings of the 2018 IEEE/RSJ International
  Conference on Intelligent Robots and Systems}, 2018, pp. 1711--1717.

\bibitem{kawaharazuka2019longtime}
K.~Kawaharazuka, K.~Tsuzuki, S.~Makino, M.~Onitsuka, Y.~Asano, K.~Okada,
  K.~Kawasaki, and M.~Inaba, ``{Long-time Self-body Image Acquisition and its
  Application to the Control of Musculoskeletal Structures},'' \emph{IEEE
  Robotics and Automation Letters}, vol.~4, no.~3, pp. 2965--2972, 2019.

\bibitem{diamond2014reaching}
A.~Diamond and O.~E. Holland, ``{Reaching control of a full-torso, modelled
  musculoskeletal robot using muscle synergies emergent under reinforcement
  learning},'' \emph{Bioinspiration \& Biomimetics}, vol.~9, no.~1, pp. 1--16,
  2014.

\bibitem{wittmeier2011caliper}
S.~Wittmeier, M.~J{\"a}ntsch, K.~Dalamagkidis, M.~Rickert, H.~G. Marques, and
  A.~Knoll, ``{CALIPER: A universal robot simulation framework for
  tendon-driven robots},'' in \emph{Proceedings of the 2011 IEEE/RSJ
  International Conference on Intelligent Robots and Systems}, 2011, pp.
  1063--1068.

\bibitem{lau2016caspr}
D.~Lau, J.~Eden, Y.~Tan, and D.~Oetomo, ``{CASPR: A comprehensive cable-robot
  analysis and simulation platform for the research of cable-driven parallel
  robots},'' in \emph{Proceedings of the 2016 IEEE/RSJ International Conference
  on Intelligent Robots and Systems}, 2016, pp. 3004--3011.

\bibitem{wittmeier2011cardsflow}
S.~Trendel, Y.~P. Chan, A.~Kharchenko, R.~Hostettler, A.~Knoll, and D.~Lau,
  ``{CARDSFlow: An End-to-End Open-Source Physics Environment for the Design,
  Simulation and Control of Musculoskeletal Robots},'' in \emph{Proceedings of
  the 2018 IEEE-RAS International Conference on Humanoid Robots}, 2018, pp.
  245--250.

\bibitem{kobayashi1998tendon}
H.~Kobayashi, K.~Hyodo, and D.~Ogane, ``{On Tendon-Driven Robotic Mechanisms
  with Redundant Tendons},'' \emph{The International Journal of Robotics
  Research}, vol.~17, no.~5, pp. 561--571, 1998.

\bibitem{niiyama2010athlete}
R.~Niiyama, S.~Nishikawa, and Y.~Kuniyoshi, ``{Athlete Robot with applied human
  muscle activation patterns for bipedal running},'' in \emph{Proceedings of
  the 2010 IEEE-RAS International Conference on Humanoid Robots}, 2010, pp.
  498--503.

\bibitem{kawaharazuka2019musashi}
K.~Kawaharazuka, S.~Makino, K.~Tsuzuki, M.~Onitsuka, Y.~Nagamatsu, K.~Shinjo,
  T.~Makabe, Y.~Asano, K.~Okada, K.~Kawasaki, and M.~Inaba, ``{Component
  Modularized Design of Musculoskeletal Humanoid Platform Musashi to
  Investigate Learning Control Systems},'' in \emph{Proceedings of the 2019
  IEEE/RSJ International Conference on Intelligent Robots and Systems}, 2019,
  pp. 7294--7301.

\bibitem{urata2006sensor}
J.~Urata, Y.~Nakanishi, A.~Miyadera, I.~Mizuuchi, T.~Yoshikai, and M.~Inaba,
  ``{A Three-Dimensional Angle Sensor for a Spherical Joint Using a Micro
  Camera},'' in \emph{Proceedings of the 2006 IEEE International Conference on
  Robotics and Automation}, 2006, pp. 4428--4430.

\bibitem{hinton2006reducing}
G.~E. Hinton and R.~R. Salakhutdinov, ``{Reducing the Dimensionality of Data
  with Neural Networks},'' \emph{Science}, vol. 313, no. 5786, pp. 504--507,
  2006.

\bibitem{kingma2015adam}
D.~P. Kingma and J.~Ba, ``{Adam: A Method for Stochastic Optimization},'' in
  \emph{Proceedings of the 3rd International Conference on Learning
  Representations}, 2015.

\bibitem{shirai2011stiffness}
T.~Shirai, J.~Urata, Y.~Nakanishi, K.~Okada, and M.~Inaba, ``{Whole body
  adapting behavior with muscle level stiffness control of tendon-driven
  multijoint robot},'' in \emph{Proceedings of the 2011 IEEE International
  Conference on Robotics and Biomimetics}, 2011, pp. 2229--2234.

\bibitem{rumelhart1986backprop}
D.~E. Rumelhart, G.~E. Hinton, and R.~J. Williams, ``{Learning representations
  by back-propagating errors},'' \emph{nature}, vol. 323, no. 6088, pp.
  533--536, 1986.

\bibitem{kawamura2016jointspace}
M.~Kawamura, S.~Ookubo, Y.~Asano, T.~Kozuki, K.~Okada, and M.~Inaba, ``{A
  Joint-Space Controller Based on Redundant Muscle Tension for Multiple DOF
  Joints in Musculoskeletal Humanoids},'' in \emph{Proceedings of the 2016
  IEEE-RAS International Conference on Humanoid Robots}, 2016, pp. 814--819.

\end{thebibliography}
}

\end{document}